\documentclass[letterpaper]{article} 
\usepackage{aaai24}  
\usepackage{times}  
\usepackage{helvet}  
\usepackage{courier}  
\usepackage[hyphens]{url}  
\usepackage{graphicx} 
\usepackage{natbib}  
\usepackage{caption} 
\usepackage{fontawesome}
\usepackage{algorithm}
\usepackage{algorithmic}
\usepackage{amsmath}
\usepackage{amssymb}
\usepackage{xcolor}
\usepackage{placeins}
\usepackage{xr}
\usepackage{arydshln}
\usepackage{pifont}
\usepackage{booktabs}
\usepackage{enumitem}
\newcommand{\cmark}{\textcolor{green}{\ding{51}}}%
\newcommand{\xmark}{\textcolor{red}{\ding{55}}}%
\newcommand{\hmark}{\textcolor{blue}{\ding{119}}}
\newcommand{\ours}{Horizon}
\usepackage{newfloat}
\usepackage{listings}
\DeclareCaptionStyle{ruled}{labelfont=normalfont,labelsep=colon,strut=off} 
\floatstyle{ruled}
\newfloat{listing}{tb}{lst}{}
\floatname{listing}{Listing}
\usepackage{multirow}
\usepackage{subcaption}

\title{HORIZON: High-Resolution Semantically Controlled Panorama Synthesis}
\author{
    Kun Yan\textsuperscript{\rm 1},
    Lei Ji\textsuperscript{\rm 2},
    Chenfei Wu\textsuperscript{\rm 2},
    Jian Liang\textsuperscript{\rm 3},
    Ming Zhou\textsuperscript{\rm 4},
    Nan Duan\textsuperscript{\rm 2},
    Shuai Ma\textsuperscript{\rm 1}
}
\affiliations{
    \textsuperscript{\rm 1}SKLSDE Lab, Beihang University,
    \textsuperscript{\rm 2}Microsoft Reseach Asia,
    \textsuperscript{\rm 3}Peking University,
    \textsuperscript{\rm 4}Langboat Technology\\
    \{kunyan,mashuai\}@buaa.edu.cn,
    \{leiji,chewu,nanduan\}@microsoft.com,
    j.liang@stu.pku.edu.cn,
    zhouming@langboat.com
}

\usepackage{xspace}
\newcommand{\supp}{supplemental material\xspace}
\DeclareMathSizes{10}{8}{5}{3.5}
\setlist[itemize]{nosep}
\begin{document}

\maketitle
\begin{abstract}
Panorama synthesis endeavors to craft captivating 360-degree visual landscapes, immersing users in the heart of virtual worlds. Nevertheless, contemporary panoramic synthesis techniques grapple with the challenge of semantically guiding the content generation process. Although recent breakthroughs in visual synthesis have unlocked the potential for semantic control in 2D flat images, a direct application of these methods to panorama synthesis yields distorted content. In this study, we unveil an innovative framework for generating high-resolution panoramas, adeptly addressing the issues of spherical distortion and edge discontinuity through sophisticated spherical modeling. Our pioneering approach empowers users with semantic control, harnessing both image and text inputs, while concurrently streamlining the generation of high-resolution panoramas using parallel decoding. We rigorously evaluate our methodology on a diverse array of indoor and outdoor datasets, establishing its superiority over recent related work, in terms of both quantitative and qualitative performance metrics. Our research elevates the controllability, efficiency, and fidelity of panorama synthesis to new levels.
\end{abstract}

\section{Introduction}

\label{sec:intro}

Panoramic images and videos are becoming increasingly popular, due to the ability to provide an unlimited field of view (FOV) compared with traditional, planar images. With panoramic images, viewers can navigate 360° views and shift the viewing perspective in all directions, capturing a wealth of environmental detail. Additionally, these images provide an immersive experience that opens up a range of possibilities for interactive applications in a variety of domains, such as advertising, entertainment, and the design industry. However, the process of panorama acquisition typically requires significant human efforts or specialized panoramic equipment. Thus, the development of automated panoramic synthesis techniques is becoming increasingly important as virtual and augmented reality technology and devices, such as head-mounted displays and glasses, continue to evolve. This technique not only helps designers save time and effort when creating and editing blueprints, but it also reduces the cost associated with specialized panoramic equipment.
\begin{figure}[hbt]
    \centering
    \includegraphics[clip, trim=4.2cm 5.7cm 5.1cm 6.6cm,width=\linewidth]{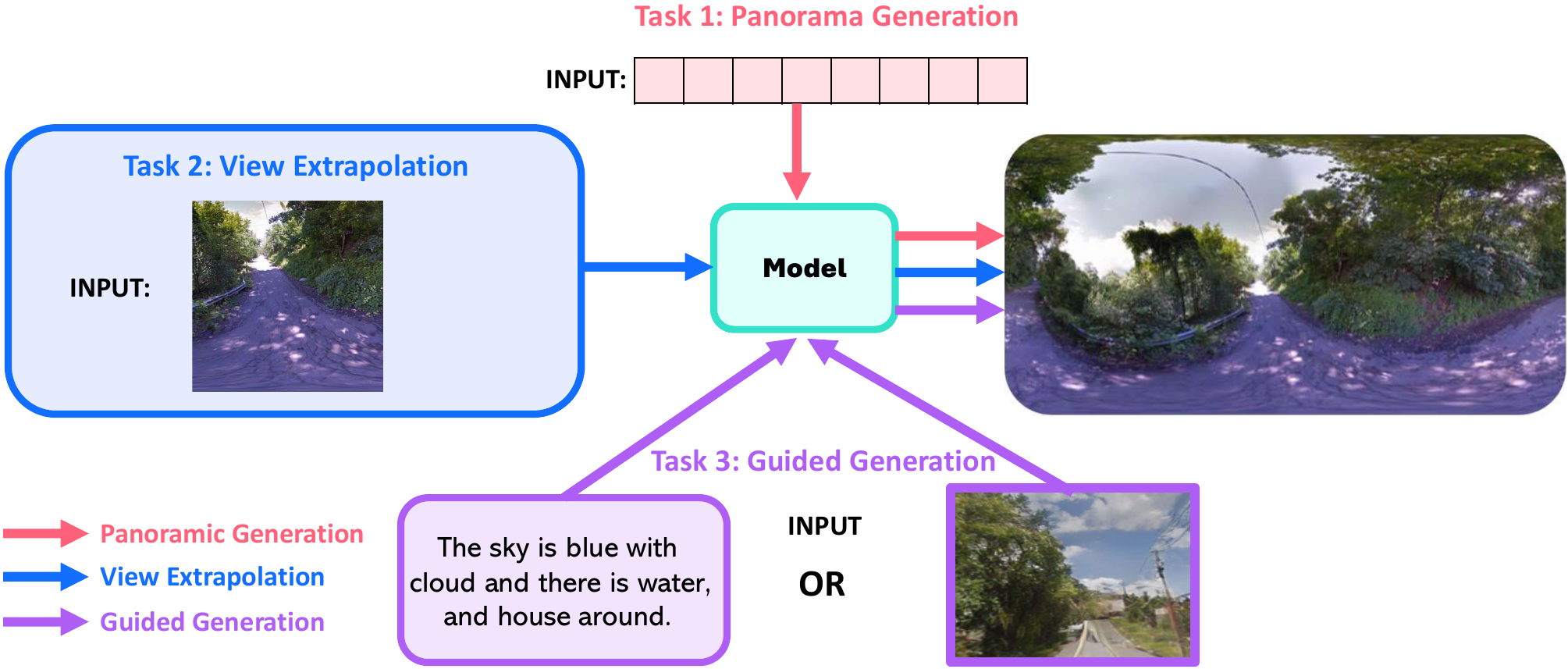}
    \caption{HORIZON supports multitype panorama synthesis}
    \label{fig:case}
\end{figure}

Panoramic images possess two unique characteristics: \emph{spherical distortion} in distinct spatial locations and \emph{continuity} between the left and right boundaries as compared to planar images. Current research on panoramic image synthesis primarily focuses on generating a spherical image with a large FOV from a single or a sequence of FOV images.\cite{sumantri2020360} introduced the use of equirectangular projection for the generation of realistic spherical images from multiple images, addressing distorted projection issues commonly encountered when working with flat images. \cite{hara2021spherical} proposed a method for generating spherical images without discontinuity. However, these methods lack the ability for \emph{user control}, which is crucial in the synthesis of virtual worlds. 

In particular, the ability to control generated content with style or semantic guidance is crucial for achieving desired images. As designers often invest a significant amount of time into creating and editing images with similar backgrounds but different semantics or styles. Research has been conducted in order to mimic human's capability to easily imagine 360-degree panoramic sceneries. This includes methods for image guidance for view extrapolation based on similar scene categories \cite{zhang2013framebreak}, scene label guidance for controlling style using a co-modulated GAN \cite{karimi2022guided}, and text guidance for image synthesis \cite{chen2022text2light}. Both view extrapolation and panoramic image synthesis tasks can benefit from a variety of inputs to guide the generation process and make this technique more flexible and useful in real-world applications. As shown in Figure \ref{fig:case}, the inputs can be text descriptions and/or visual inputs for semantic and style guidance.

It is worth noting that recent text-to-image generation methods, such as VAE~\cite{DALLE}, GAN~\cite{esser2021taming}, and Diffusion~\cite{dhariwal2021diffusion}, have achieved great success in terms of semantic relevance and controllability for planar images. These works leverage the capabilities of attention mechanism~\cite{vaswani2017attention} to take various types of guidance as generating conditions. However, these methods are difficult to apply directly to panoramic image synthesis due to a lack of consideration for spherical characteristics. To the best of our knowledge, there are currently no existing frameworks that are general enough to handle all of these spherical properties and controlling guidance for panorama synthesis in a unified framework. This is the motivation behind our work, which aims to explore advanced guided image generation techniques and design specific modules, mechanisms, and training strategies for spherical structures.

Additionally, it is imperative to generate \emph{high-resolution} panoramic images in order to enhance the immersive experience for Virtual Reality and Augmented Reality. However, simply adopting current state-of-the-art methods such as DALL·E-like~\cite{esser2021taming} for high-resolution panorama generation can result in spherical distortion and low efficiency. Recent efforts have attempted to address this issue by using separate models to first generate low-resolution panoramas and then upscaling to high-resolution, such as \cite{akimoto2022diverse,chen2022text2light}. These methods, however, can still result in artifacts caused by error accumulation. It is worth noting that we find properly managing high-resolution context within a defined module can alleviate this problem and produce better results.

In this paper, we introduce a novel, versatile framework for generating high-resolution panoramic images that have well-preserved spherical structure and easy-to-use semantic controllability. Specifically, we employ a two-stage procedure that includes learning an image encoder and decoder in the first step and a reconstruction model in the second step. In the second stage, we propose a new method called Spherical Parallel Modeling(SPM) that not only improves efficiency through parallel decoding, but also addresses the issue of local distortion through the use of spherical relative embedding and spherical conditioning improvement. Additionally, we have found that the panoramic pictures generated by SPM no longer have the problem of screen tearing when the left and right edges are spliced, which means that the generated panoramic pictures can be directly viewed in VR devices without the need for further editing.

The contributions can be summarized as:
\begin{itemize}[noitemsep]
\item Alleviating the spherical distortion and edge in-continuity problem through spherical modeling.
\item Supporting semantic control through both image and text guidance.
\item Effectively generating high-resolution panoramas through parallel decoding. 
\end{itemize}

\section{Related Work}
\label{sec:related}
\paragraph{Panorama Synthesis}
\begin{table}[]
    \centering
    \resizebox{\linewidth}{!}{
    \begin{tabular}{cccccc}
    \toprule
    \multirow{2}{*}{\shortstack{Method}} & \multirow{2}{*}{\shortstack{ High\\ Resolution?}} & \multirow{2}{*}{\shortstack{Spherical \\ Coherence?}} & \multirow{2}{*}{\shortstack{Global  \\Semantic Condition?}} & \multirow{2}{*}{\shortstack{Multiscale \\Semantic Editing?}} & \multirow{2}{*}{\shortstack{Inference \\ Efficiency?}} \\
    & \\
    \midrule
    COCO-GAN~\shortcite{lin2019cocogan} & \xmark & \hmark & \hmark & \xmark &\faRocket\\
    InfinityGAN~\shortcite{lin2021infinity} & \cmark & \xmark & \cmark & \cmark&\faRocket\\
    LDM~\shortcite{rombach2021highresolution} & \cmark & \xmark & \cmark & \xmark & \faMale \\
    Text2light~\shortcite{chen2022text2light} & \cmark & \hmark & \hmark & \xmark & \faMale\\
    OminiDreamer~\shortcite{Omnidreamer} & \xmark & \hmark & \xmark & \xmark &\faCar \\
    \midrule
    \ours & \cmark & \cmark & \cmark & \cmark & \faRocket\\
    \bottomrule    
    \end{tabular}
    }
    \caption{The comparison between our method and the most recent relevant methods. \hmark~represents partially satisfying the property. Find detailed evaluations and explanations in the appendix.}
    \label{tab:comparision}
\end{table}

Panorama synthesis, a well-established task in computer vision, involves various input types such as overlapped image sequences \cite{szeliski2006image,brown2007automatic}, sparse images \cite{sumantri2020360}, and single images \cite{akimoto2019360,hara2021spherical}. Traditional methods employed image matching and stitching \cite{szeliski2006image}, while recent generative models utilize GAN-based methods \cite{akimoto2019360,koh2022simple} and autoregressive models \cite{rockwell2021pixelsynth} for panorama generation.

In computer graphics, view synthesis techniques, including geometry and layout prediction, optical flow, depth, and illumination estimation \cite{song2019neural,xu2021layout,zhang2022generating,wang2022stylelight,somanath2021hdr}, are often studied. Spherical structure and texture are modeled using cube maps \cite{han2020piinet}, cylinder convolution \cite{liao2022cylin}, predicted panoramic three-dimensional structures \cite{song2018im2pano3d}, and scene symmetry \cite{hara2021spherical}.Most previous generative models are limited in handling fixed scenes and low-resolution images. Recent research \cite{sumantri2020360,akimoto2022diverse} addresses these limitations using hierarchical synthesis networks \cite{sumantri2020360}, U-Net structures \cite{akimoto2022diverse}, and separate models for generating and upscaling low-resolution images \cite{chen2022text2light}. Our proposed method tackles high-resolution panorama context and preserves spherical structure within a single module.

User-controlled semantic content generation is essential for interactive panorama generation. Recent works adopt scene symmetry with CVAE-based methods \cite{hara2021spherical} or scene category with GAN-based methods \cite{karimi2022guided} for view extrapolation. Our versatile framework addresses spherical, user-controlling, and high-resolution panoramas in a unified manner, incorporating mechanisms to handle spherical distortion, continuity, and semantic guidance without additional tuning.

We put a straightforward comparison of these most recent relevant efforts in Table~\ref{tab:comparision}, and detailed discussion can be found in the appendix.
\paragraph{Image Generation}
Previous image generation works primarily employ generative adversarial networks (GAN)~\cite{goodfellow2014generative,reed2016generative,Xu2018AttnGANFT,qiao2019mirrorgan,XMCGAN}, VAE-based methods~\cite{kingma2013auto,NIPS2017_7a98af17}, and denoising diffusion models~\cite{nichol2021glide,gu2021vector,kim2021diffusionclip}. With the advent of transformer models, two-stage methods have emerged as a new paradigm for pretraining with web-scale image and text pairs, demonstrating effectiveness in generalizing high semantically related open-domain images~\cite{DALLE,ding2021cogview,zhang2021m6}. These methods tokenize images into discrete tokens using VQVAE~\cite{NIPS2017_7a98af17} or VAGAN~\cite{esser2021taming}, then generate visual tokens for decoding into real images.

Efforts have been made to improve high-resolution planar image generation~\cite{esser2021taming, Chang2022MaskGITMG, nuwa-infinity}. However, these models often produce blurred or teared artifacts, making them unsuitable for panoramic scenarios. Guided image generation methods achieve superior performance~\cite{DALLE,ding2021cogview,zhang2021m6}, but applying existing models to panoramic generation without considering the unique spherical structure remains challenging.

\section{Method}
\label{sec:method}
The overall training is a two-stage procedure, similar to \cite{DALLE,ding2021cogview}. The first stage is to train an encoder for image/view representation (discrete visual tokens in this paper) and a decoder for image generation, both of which are frozen in the second stage. The second stage is to learn a reconstruction model based on the discrete visual representation.

\begin{itemize}[noitemsep]
    \item \textbf{Stage 1.} Every equirectangular projected panoramic image with a resolution of 768x1,536 is first divided into 3x6=18 RGB view patches, each with a resolution of 256x256. Then we train a VQGAN\cite{esser2021taming} on every view patch separately. The encoder of VQGAN compresses each RGB view patch into a 16×16 grid of view tokens. The overall view token dictionary has a size of 16,384 possible values. As a result, each panoramic image has 18×16×16 view tokens. 
    \item  \textbf{Stage 2.} All 18 groups of view tokens from a single panoramic image are modeled as a whole context to incrementally learn the reconstruction of all view tokens. We progressively develop auto-regressive modeling, local parallel modeling, and the newly proposed spherical parallel modeling detailed described in the following subsections. 
\end{itemize}
We apply the off-the-shelf model in the first stage and mainly devote our effort to effectively modeling the prior in the second stage. Due to the large number of view tokens for a single panorama (18x16x16=4608), it is still a non-trivial problem to learn the reconstruction. We should balance the quality, efficiency, and controllability with considerable refinement. In the following subsections, we will describe our progressive attempts and corresponding design considerations.

\subsection{Auto-Regressive Modeling }
One intuitive way is to directly employ a auto-regressive transformer decoder to generate 4608 view tokens one by one. However, due to the quadraticity of the attention mechanism of the transformer itself, directly inputting 4608 tokens into the model will bring huge memory consumption and great difficulties to the training of the model. At the same time, the sequence is too long for the model to converge efficiently.

\begin{figure*}[t]
\centering
  \subfloat[ARM and LPM]{
    \includegraphics[clip, trim=1.5cm 2cm 3cm 2.5cm,width=0.45\textwidth]{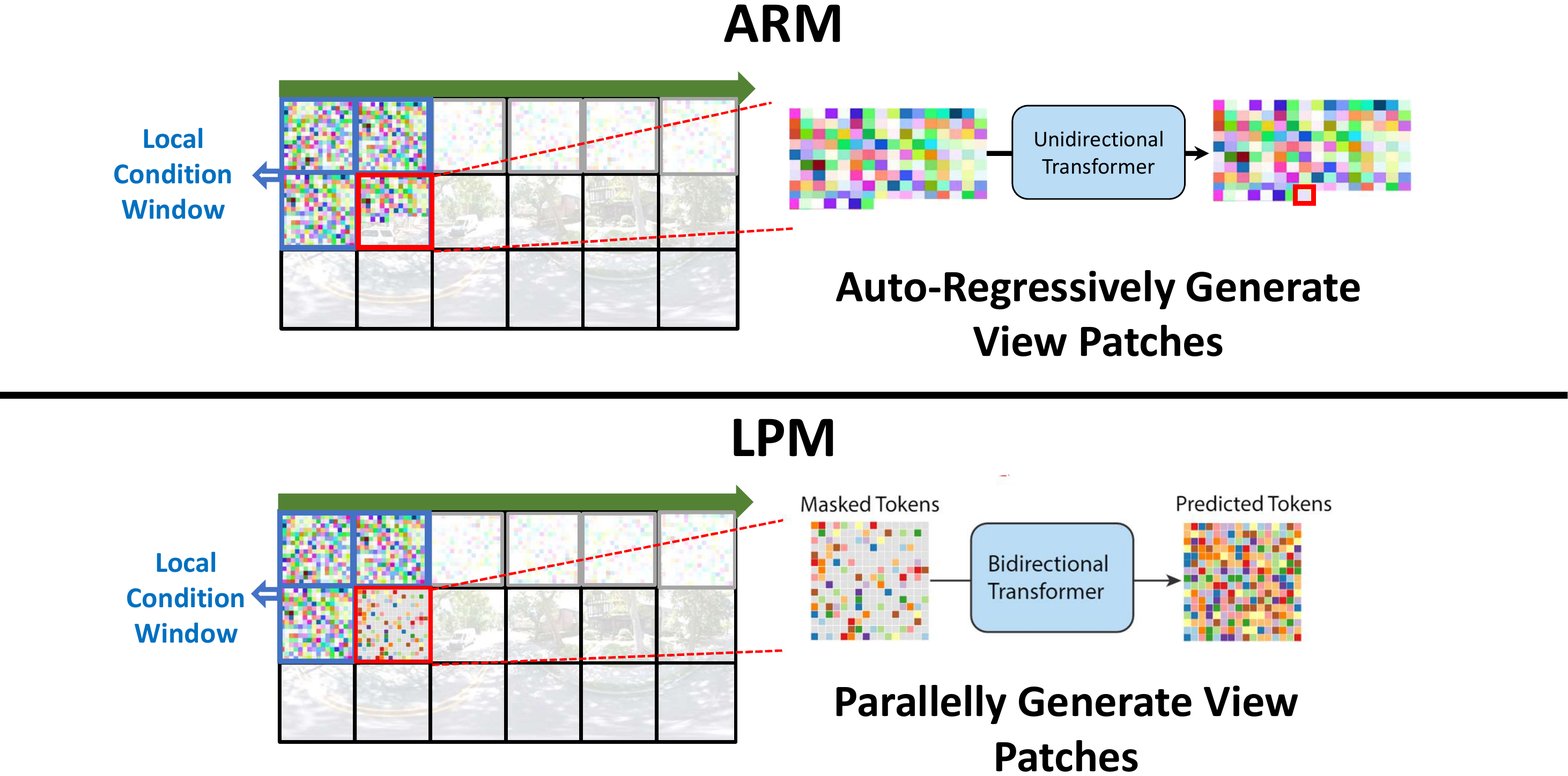}
  }
  \subfloat[SPM]{
    \includegraphics[clip, trim=3cm 2cm 3cm 2cm,width=0.45\textwidth]{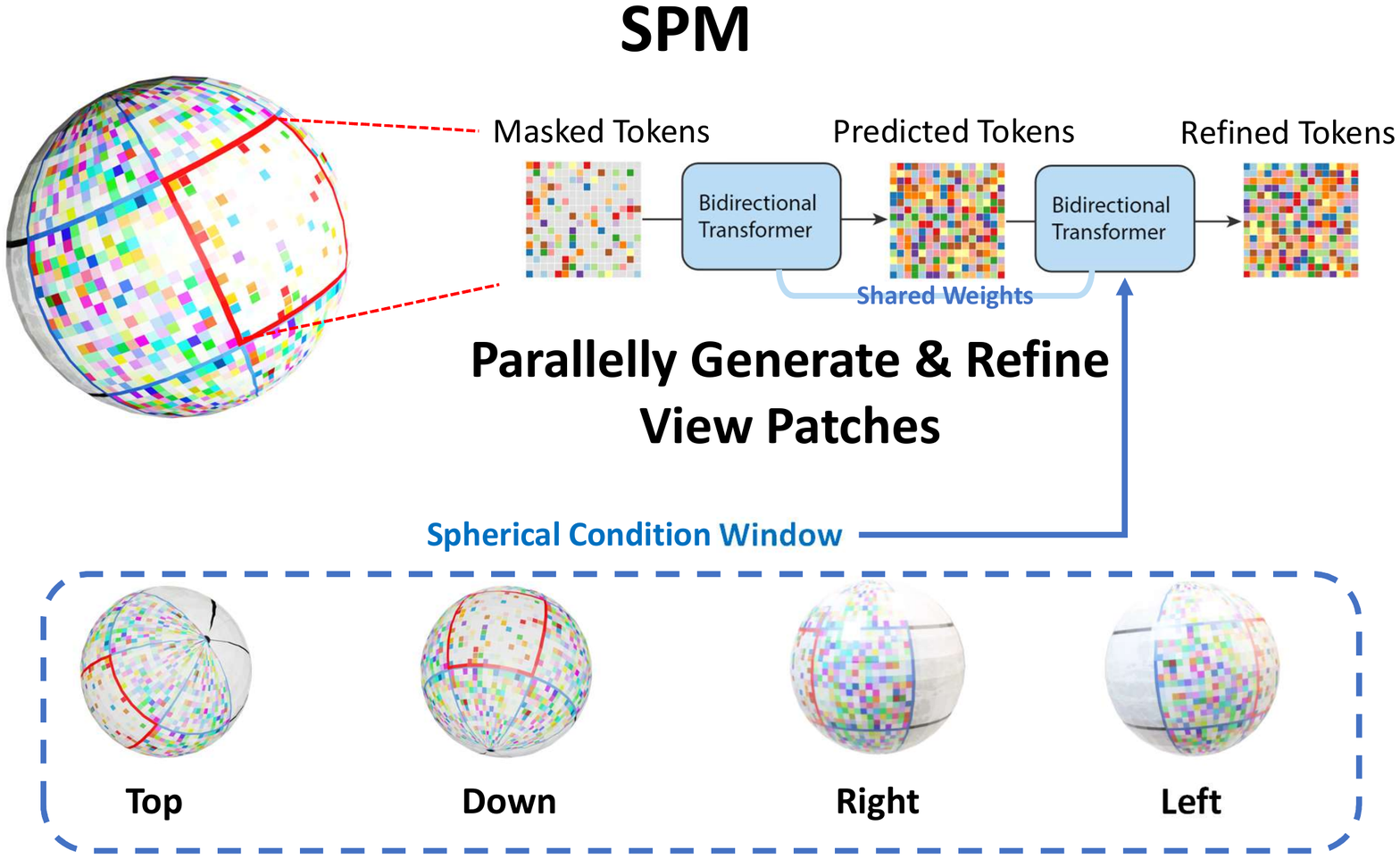}
  }
   \caption{Modeling Strategy: we progressively improve modeling strategy from ARM to LPM and eventually SPM, achieving both high efficiency and high fidelity. In this Figure, the red boxes 
   show the current view patch to be generated, while the blue boxes present the condition window.
   }
   \label{fig:2stage}

\end{figure*}

We noticed that, as the relative distance increases, the impact of adjacent tokens becomes weak or even negative for the quality. Therefore, we shrink the attention scope for both efficiency and effectiveness consideration. Specifically, the range of attention for each view patch is limited to 2 surrounding view patches on the left and above, and autoregressively performs the prediction within the current patch. The interval of attention is shown at the top of Figure \ref{fig:2stage}.
After making this improvement, we have been able to generate decent high-resolution panoramas, which are also used as our first baseline ARM.

\subsection{Local Parallel Modeling}
Although ARM makes high-resolution panorama generation basically feasible, flattening the view patch into a one-dimensional sequence of tokens in raster scan order is still not an optimal and efficient modeling solution. Since the length of the autoregressive sequence still grows quadratically, it not only presents a challenge for modeling long-term correlations but also makes decoding intractable. Inspired by MaskGIT\cite{Chang2022MaskGITMG}, we adopt the Masked Visual Token Modeling(MVTM) into the view modeling process, which can be formulated as:
\begin{equation}
    \mathcal{L}_{\text{LPM}}=-{\mathbb{E}}\left[\sum_{\forall i \in[1, N], mask_{i}=1} \log p\left(y_{i} \mid Y_{\overline{\mathbf{M}}}; Y_W\right)\right]
\end{equation}
For every training pass, sample a subset of tokens and replace them with a special [MASK] token. The number of masked tokens is parameterized by a 
scheduling function $\lceil\cos(r*\pi /2) \cdot N\rceil$, $r$ is a real number uniformly sampled from 0 to 1, N is the total number of view tokens in current view patch.
Masked token sequence $Y_{\overline{\mathbf{M}}}$ and ground truth view tokens from surrounding view patch $Y_W$ are fed into a multi-layer bidirectional transformer to predict the probabilities $p\left(y_{i} \mid Y_{\overline{\mathbf{M}}}; Y_W\right)$ for each masked token, where the negative log-likelihood is computed as the cross-entropy between the ground-truth and prediction.

During inference, we use a similar iterative decoding technique with a constant step $T$, initially all tokens in the current patch are masked, at each step $t$ only $\left\lceil(1-(\cos\left(\frac{\pi t}{2T}\right))) N\right\rceil$ tokens with higher confidence are kept, others will be refined during further steps.
We named this adapted version of MaskGIT as Local Parallel Modeling (LPM), that is, the modeling is applied in parallel for each local view patch. By applying this strategy, we shorten the inference speed by 64 times. However, we also observe a significant performance drop compared with ARM. 

\subsection{Spherical Parallel Modeling}
Both ARM and LPM modules regard each patch view equally in the image, which do not take the spherical characteristics of panoramic images into consideration in model design. They assume that the visual features after spherical projection are translation-invariant on the two-dimensional plane. This is obviously not in line with the actual situation. We observed that under the ARM method, the model can still maintain a strong relative positional relationship, and according to the current sequence order, it can be deduced what degree of deformation should be used to generate the current view token. However, under the LPM method, the generation of the current token is no longer strictly constrained by the previous token, and sequence order is no longer an important factor for model learning goals. Naturally, the panoramic images generated by LPM have distorted local details, which are relatively weak in performance indicators such as FID. 

In this section, we describe a new Spherical Parallel Modeling(\textbf{SPM}) method that not only maintains the efficiency of parallel decoding but also alleviates the local distortion problem through the spherical relative embedding and spherical conditioning improvement. Besides, we found that the panoramic images generated by SPM no longer have the problem of screen tearing when the left and right edges are spliced.

\subsubsection{Spherical Relative Embedding}

Relative positional embedding effectively captures positional information during attention, particularly for spherical properties. Formally, a positional encoding function $f(\mathbf{x},l)$ is defined for item $x$ at position $l$. For items $\mathbf{q}$ and $\mathbf{k}$ at positions $m$ and $n$, the inner product between $f(\mathbf{q},m)$ and $f(\mathbf{k},n)$ depends on $\mathbf{q}$, $\mathbf{k}$, and their relative position $m-n$. The dot product between two vectors is a function of their magnitudes and the angle between them.

Rotary Position Embedding (RoPE) \cite{rope-paper} encodes text token embedding with an absolute position using a rotation matrix, incorporating explicit relative position dependency in self-attention. Embeddings are treated as complex numbers and positions as pure rotations. During attention, if both query and key are shifted by the same amount, changing the absolute but not relative position, both representations are rotated similarly, maintaining the angle and dot product between them.

The function solution that satisfies the above requirement can be formulated as below:
\begin{equation}
\begin{aligned}
f(\mathbf{q}, m) & = \begin{pmatrix}
M_1 & & & \\
& M_2 & & \\
& & \ddots & \\
& & & M_{d/2}
\end{pmatrix}
\begin{pmatrix}
q_1\\
q_2\\
\vdots\\
q_d
\end{pmatrix} 
 & =   
\mathbf{R_m Q_m}\\
& = \mathbf{R_m W_q X_m}
\end{aligned}
\end{equation}
where $ M_j=\begin{pmatrix}\cos m\theta_j & -\sin m\theta_j \\sin m\theta_j & \cos m\theta_j\end{pmatrix}$, $\mathbf{R_m}$ is the block diagonal rotation matrix, $\mathbf{W_q}$ is the learned query weights, and 
$\mathbf{X_m}$ is the embedding of the $m$-th token. For query $\mathbf{k}$, a similar corresponding equation is applied. When extending to the 2-dimensional case, the rotation matrix should correlate with both coordinates $x$ and $y$:

\begin{equation}\boldsymbol{M}_{x,y}=\left( 
\begin{array}{cc:cc} 
\cos x\theta & -\sin x\theta & 0 & 0 \\ 
\sin x\theta & \cos x\theta & 0 & 0 \\ 
\hdashline 
0 & 0 & \cos y\theta & -\sin y\theta \\ 
0 & 0 & \sin y\theta & \cos y\theta \\ 
\end{array}\right)\label{eq:rope-2d}\end{equation}

However, only two-dimensional relative position embedding cannot represent the relative positional relationship of the spherical surface. This is manifested in two aspects. One is that the distance between the plane and the spherical surface is measured in different ways. Second, the coordinates of the same latitude have a ring-shaped positional relationship, that is, for a token sequence $0 \dots m$ at the same latitude, the positional embedding of token $0$ and the positional embedding of token $m$ should be as close as possible. To satisfy this property, we re-derived the rotation matrix, instead of the $\Theta$ in the original RoPE:
\begin{equation}
    \Theta=\left\{\theta_{i}=10000^{-2(i-1) / d}, i \in[1,2, \ldots, d / 2]\right\}
\end{equation}
We define $\Theta_{sphere}$ as:
\begin{equation}
    \Theta_{shpere, x}=\left\{\theta_{i}={\frac{-2(i-1)*2\pi}{d * w} }, i \in[1,2, \ldots, d / 2]\right\},
\end{equation}
\begin{equation}
    \Theta_{shpere, y}=\left\{\theta_{i}={\frac{-2(i-1)*\pi}{d * h} }, i \in[1,2, \ldots, d / 2]\right\},
\end{equation}
where $x,y$ are the different axis of the spherical surface, $w$ is the length of token sequences along the $x$ axis, and $h$ is the length along the $y$ axis. Note that as $\Theta_{shpere, x}$ represents latitude, the numerator has a factor $2\pi$, which makes the rotation of the sequence head and tail as close as possible. While $\Theta_{shpere, y}$ does not keep this property, as the poles of a sphere are far from each other naturally.

Spherical Relative Embedding(SRE) applies to self-attention as follows:
\begin{equation}
\begin{aligned}
        SRE(\boldsymbol{q}_{(x1,y1)}^{\boldsymbol{\top}})SRE( \boldsymbol{k}_{(x2,y2)})    =\left(\boldsymbol{R}_{\Theta, (x1,y1)}^{d} \boldsymbol{W}_{q} \boldsymbol{x}_{(x1,y1)}\right)^{\boldsymbol{\top}} \\ \left(\boldsymbol{R}_{\Theta,(x2,y2)}^{d} \boldsymbol{W}_{k} \boldsymbol{x}_{(x2,y2)}\right)
\end{aligned}
\end{equation}

\subsubsection{Spherical Conditioning}

The autoregressive one-directional transformer decoder generates the tokens from left to right, which leads to discontinuity between the left-most and right-most boundaries. When viewing these images with the panorama viewer, we see obvious tearing artifacts at the stitching seams. To make the left-most and right-most pixels consistent, these pixels should be generated with consideration of each other. This local detail also implies a deeper defect of the aforementioned method. If the context information of the complete spherical structure is not considered when predicting the token of the current position, the consistency and integrity of the final overall result cannot be guaranteed.

In order to fix this defect, we redesign the conditions for generating each view patch. Through the two-pass mechanism, the model no longer only autoregressively focuses on the small window on the upper left but also focuses on the entire hemispherical area around the current patch view. 
Specifically, in the training phase, the model learns both $ \mathcal{L}_{\text {LPM}}$ and $ \mathcal{L}_{\text {SPM }}$ in each iteration step. 
The form of $ \mathcal{L}_{\text {SPM }}$ is as follows:
\begin{equation}
    \mathcal{L}_{\text {SPM }}=-{\mathbb{E}}\left[\sum_{\forall i \in[1, N], mask_{i}=1} \log p\left(y_{i} \mid Y_{\overline{\mathbf{M}}}; Y_S\right)\right]
\end{equation}
It is worth noting that $Y_S$ is different from $Y_W$ in $ \mathcal{L}_{\text {LPM}}$. For the view patch at row $i$ and column $j$, the corresponding $Y_W$ contains view patches of upper and left, while $Y_S$ contains these of upper, down, left, and right. Specifically, $Y_S$ comprises of $[(i-1, j), (i, j-1), (i-1,j-1)]$, and $Y_S$ comprises of $[(i-1, j-1), (i-1,j), (i-1, j+1), (i,j-1),(i,j+1),(i+1,j-1),(i+1,j),(i+1,j+1)]$. When the above coordinates are out of bounds, $Y_W$ will not consider the part beyond the boundary, and $Y_S$ will extend the part beyond the x-axis to the other side. In addition, each $Y_S$ also applies the spherical relative embedding transformation described above and a special learnable phase embedding $\mathbf{I}$ to let the model know which pass it is currently in.

During inference, the model first performs a complete LPM decoding and retains all tokens, and then superimposes the spherical relative embedding and phase embedding $\mathbf{I}$ for each token in the second pass to optimize the generation result of the first pass. In this way, although the inference time is doubled, it is still faster than ARM and can further greatly improve the generation quality, surpassing both ARM and LPM.

\subsubsection{Guided Semantic-Condition} 
To further enhance the capabilities of the HORIZON model, we employ semantics as an additional input condition to guide the panoramic image generation process. Specifically, we use FOV=90 degrees to cut out the front, back, left, and right perspective pictures of the panoramic image and encode them with the pretrained CLIP\cite{radford2021learning} visual module to get four semantic vectors for each panorama. We then take those vectors as semantic conditions sequentially appended to sphere conditions, and train in an end-to-end way. Similarly, we also employ the pretrained CLIP text module to encode text as a semantic condition to control the generation. During inference, we can either input visual conditions or text conditions respectively.

\section{Experiment Setup}
\subsection{Dataset}

We evaluate our model on the high-resolution StreetLearn dataset \cite{mirowski2019streetlearn}, which consists of Google Street View panoramas. We use the Pittsburgh dataset containing 58k images, split into 52.2k for training and 5.8k for testing. The equirectangular panorama RGB images are stored as high-quality JPEGs with dimensions of 1664 x 832. In our experiments, we resize all panoramas to 1536 x 768. The experiments are conducted on 64 V100 GPUs, each with 32GiB memory. Our model is evaluated on three typical panorama tasks: panorama generation, view extrapolation, and guided generation.

To further show the flexibility of our method on arbitrary resolutions, we also conducted experiments on a higher resolution setting of 3072x1536. However, as most previous works are unable to handle such large images, we present only qualitative demonstrations in the appendix.To show our method are applicable to diverse scenes, we conduct experiments on  Matterport 3D with different available baselines\cite{chen2022text2light,lin2019cocogan,lin2021infinity}. Please find it in the appendix.

\subsection{Task 1: Panorama Generation}

\begin{table*}[t]
  \centering
    \begin{tabular}{l|c|cccc|c}
    \hline
     & FID$\downarrow$ &  \multicolumn{4}{c|}{Spherical FID$\downarrow$} & Continuity \\
                  &  & mean&top&middle&bottom&LRCS$\downarrow$\\
    \hline        Text2light\cite{chen2022text2light}&36.33&  56.31&48.05&60.28&60.62&   0.0224\\
    ARM&25.36&  41.21&26.16&32.17&65.32&   0.0283 \\
    LPM&45.71 & 57.13 &64.34&46.71&60.35&  0.2032\\
    \hline
    SPM(+SRE)&10.74& 39.18&28.72&26.44&62.39& 0.0726 \\
        SPM(+SRE+SC) &\textbf{7.79}&  \textbf{20.97}&\textbf{15.82}&\textbf{21.80}&\textbf{25.29}&\textbf{0.0020} \\
    \hline
    \end{tabular}
  \caption{ Generation results. SPM is the spherical model in our paper, SRE is spherical relative embedding and SC is spherical conditioning.}
\label{tab:generation}
\end{table*}

The widely used Frechet Inception Distance (FID) score \cite{FID} evaluates image quality by measuring feature distribution distances between real and fake images. However, FID treats all image positions equally, while information density in spherical images varies spatially. The top and bottom of an image often contain sparse information, while the middle holds denser information.

To account for this, we propose a novel \emph{spherical FID} for evaluating panoramic image quality, which dynamically considers information density variation. FID is calculated on different view patch sets within an image. For a panorama with a 3-row and 6-column grid of view patches, we label each row as top, middle, and bottom, and calculate FID scores for these subsets. Table \ref{tab:generation} reports spherical FID scores for various locations. As a baseline, we train the text2light model \cite{chen2022text2light} on the Streetlearn dataset, with detailed settings in the appendix.

Our SPM(+SRE+SC) model generates state-of-the-art results, significantly outperforming baseline models. Both spherical relative embedding and spherical conditioning are effective mechanisms.The second pass balances efficiency and effectiveness by revising the first pass. LPM is the most efficient model, albeit with lower performance. 

\begin{figure}[htb]
    \centering
    \includegraphics[clip, trim=0cm 3.2cm 0cm 3.2cm,width=0.5\textwidth]{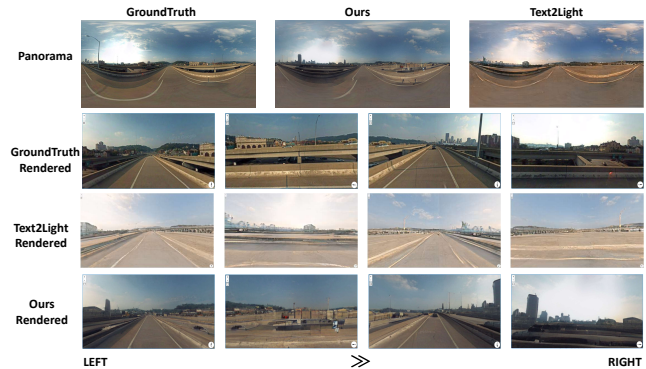}
    \caption{Generated examples. The first row presents the ground-truth 
    and generated images. The second to the fourth rows are the four snapshots of rendered results of each methods respectively by using the 360 panorama viewers. }
    \label{fig:allcases}
\end{figure}

We present several examples generated in Figure \ref{fig:allcases}. We rely on the open tool Online 360° Panorama Viewer VR \footnote{https://renderstuff.com/tools/360-panorama-web-viewer/} to view the image. For each case, we present the snapshot of the four views from left to right in the Figure. More cases are listed in the 
\supp.

\begin{figure}[h!]
    \centering
    \includegraphics[width=0.95\linewidth]{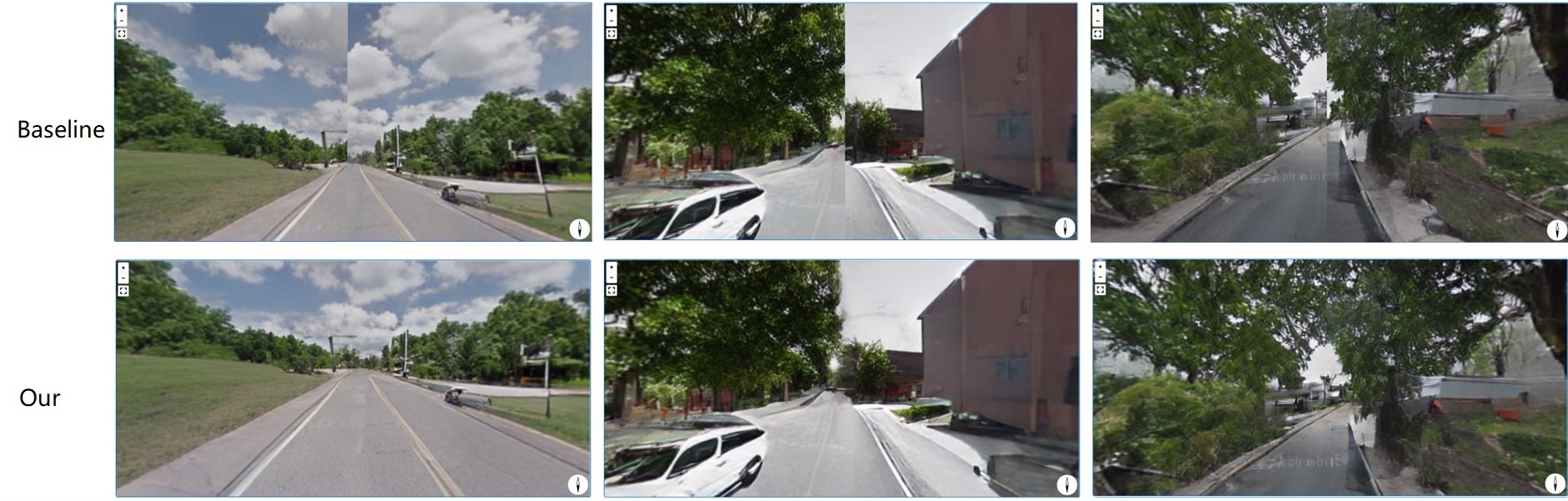}
    \caption{Discontinuity v.s. Continuity. The three randomly selected cases present results from baseline and our models. The top images are the generated images of the baseline(LPM) method and the bottom examples are from our model(SPM). There is an obvious split line in the middle of each image on the top examples while the boundary is smooth on the bottom examples.}
    \label{fig:continuty}

\end{figure}

\paragraph{View Discontinuity Problem}
The discontinuity of synthesis occurs when the left-most and the right-most boundary merged into one spherical image. Among the four snapshots from the viewer tool, the last (4th) image rendered is the merged image in which the middle is exactly the boundary between the left most and right most. From the showcases in Figure \ref{fig:continuty}, we can see that the results of the baseline algorithm have an obvious separator in the middle while our algorithm considering the spherical attention generate a smooth connection. Moreover, we evaluate this continuity quantitatively by gradient based metrics as shown in Table \ref{tab:generation}. Inspired by the metric Left-Right Consistency Error (LRCE)  \cite{shen2022panoformer} for depth estimation, we evaluate the consistency of the left-right boundaries by calculating the horizontal gradient between the both sides of the panorama. In details, the horizontal gradient $G^H_I$ of the image I can be written as $G_I = \max\limits_{dim=-1}|I^{col}_{first} - I^{col}_{last}|$, where $I^{col}_{first}/ I^{col}_{last}$ represents the RGB values in the first/last columns of the image $I$. 
Note that, different from LRCE, the generated panorama can not minus ground truth gradient to alleviate natural discontinuity. We choose to calculate the distribution distance instead of the absolute distance between the predicted horizontal gradient and real panorama gradient to measure the boundary continuity. The final calculation of LRCS(left-right continuity score) is as follows:
\begin{equation}
    LRCS = KL(\mathcal{N_{P}},\mathcal{N_{GT}})
\end{equation}
$\mathcal{N_{P}}$ and $\mathcal{N_{GT}}$ are two normal distributions estimated from the horizontal gradient of the predicted panorama($\{G_P\}$) and ground truth($\{G_{GT}\}$), respectively. $KL$ means KL-distance. The lower LRCS means the panorama is more seamless. We show the result in Table \ref{tab:generation}, though Parallel Decoding increase the discontinuity, after using SRE and SC our final results has significantly resolve the problem and achieves 10 times lower LRCS than Text2light.
All the above results demonstrates the effectiveness of our proposed spherical attention module. 

\subsection{Task 2: View Extrapolation} 
We conduct quantitative experiments and adopt structural similarity (SSIM) and peak-to-signal-noise-ratio (PSNR) and FID as evaluation metrics specific for the view extrapolation tasks. 

Our generative model demonstrates superior performance compared to baseline methods as demonstrated in Table \ref{tab:extrapolation}. To further illustrate the effectiveness of our approach, we have included a comparison of our method with Omnidream \cite{akimoto2022diverse} in the appendix, where we have constrained the resolution to 1024x512 in accordance with their capabilities.

\begin{table}[t]
  \centering
    \begin{tabular}{l|rr|r} 
    \hline
     &  $SSIM\uparrow$ & $PSNR\uparrow$ & FID$\downarrow$ \\ 
    \hline
    ARM &  0.521&15.39 & 11.78 \\ 
    LPM &  0.508&14.94 & 17.62\\ 
    \hline
    SPM &    0.542&15.49 &5.53\\ 
    \hline
    \end{tabular}
  \caption{ View Extrapolation Results. 
  }
  \label{tab:extrapolation}
\end{table}

\begin{figure}[h!]
    \centering
    \includegraphics[clip, trim=3cm 4.5cm 2.5cm 4.3cm,width=0.45\textwidth]{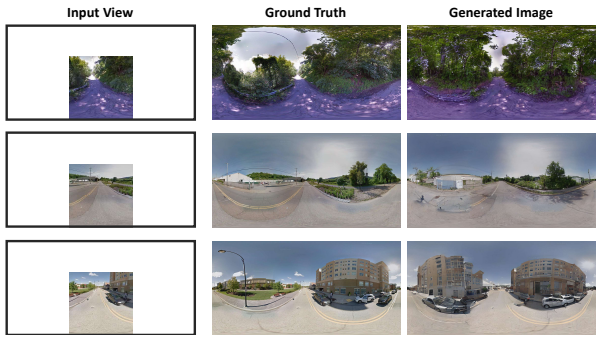}
    \caption{ View Extrapolation. The first column gives the input samples, the second column presents the ground truth examples, and the third column demonstrates the generated panorama.}
    \label{fig:extrapolation}
\end{figure}

Our generative model demonstrated exceptional performance in view extrapolation, as validated by the examples shown in Figure \ref{fig:extrapolation}. The generated panoramas not only seamlessly filled in unseen content, but also possessed reasonable structure and rich semantics. These results showcase the superior capabilities of our model in generating high-resolution, coherent panoramas.

\subsection{Task 3: Guided Generation}
\begin{figure}[h!]
    \centering
    \includegraphics[clip, trim=0.5cm 4.5cm 0.5cm 4.3cm,width=0.48\textwidth]{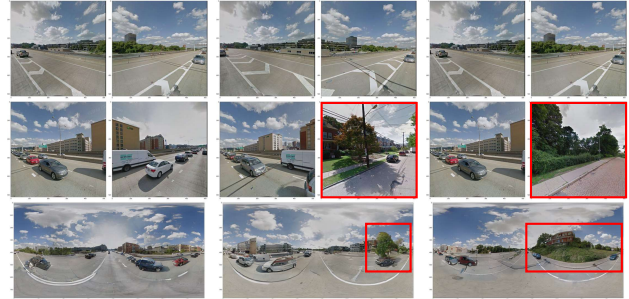}
    \caption{ Visual Guided generation. When we replace the guidance with different visual semantics as shown in the middle and right columns, we can manipulate the generated panoramas as we need.}
    \label{fig:control}
\end{figure}

\begin{figure}[h!]

    \centering
        \includegraphics[clip, trim=1.8cm 4.5cm 0.8cm 4.3cm,width=0.5\textwidth]{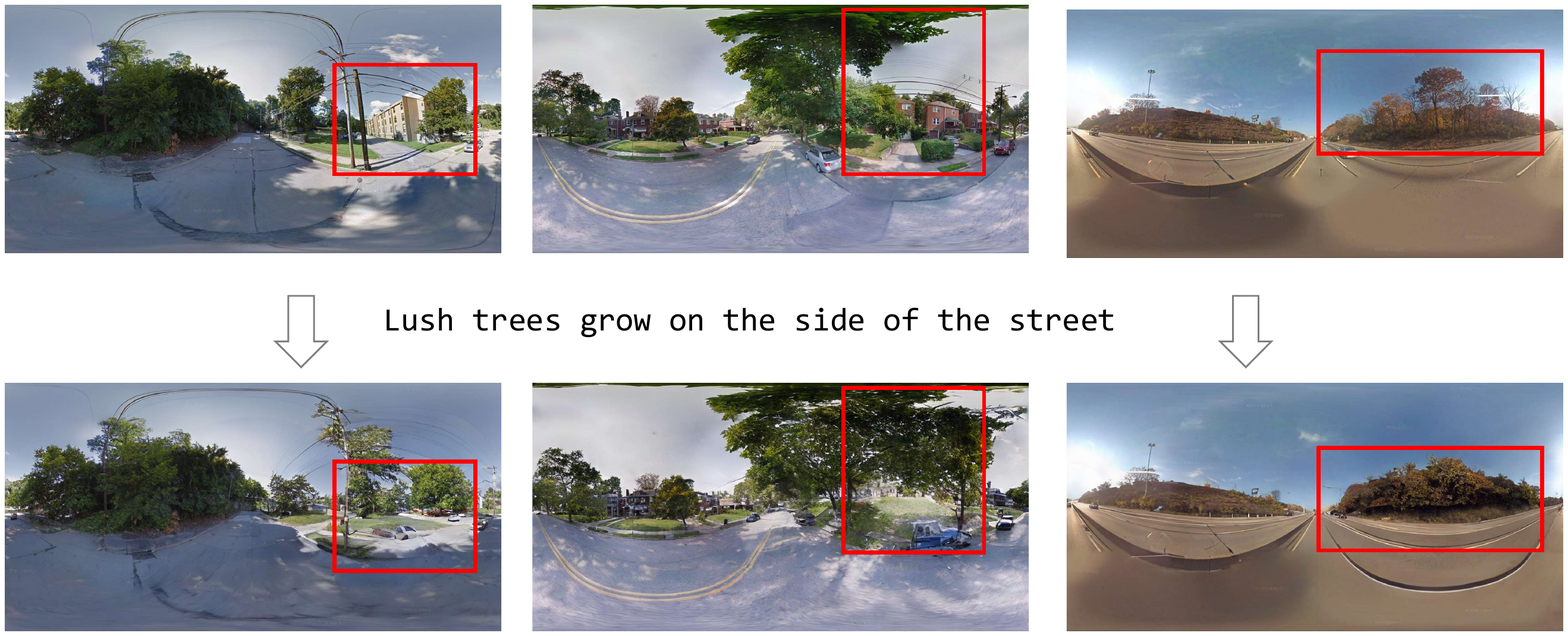}
      \\
    \caption{ Textual Guided generation. We can also use natural language to edit or embellish target panoramas. In each case, the top role are the original panoramas, and the bottom role are the embellished panoramas according to the text hint shows in the middle.}
    \label{fig:control-text}
\end{figure}

We illustrate the guided generation showcases in Figure \ref{fig:control} and Figure \ref{fig:control-text}. The visual guidance and text guidance are encoded by CLIP model. From these examples, we can observe that our framework can edit and modify semantic elements in the panorama by providing reference view images or text hints at specific locations. More cases can be found in the \supp. 

\paragraph{Visual Guidance} 
As shown in Figure \ref{fig:control}, 
the case presents the results given the visual guidance. The left case shows the original panoramic image (bottom) as well as 4 FOV images rendered (top 4 images). The middle and right cases demonstrate the generated results given the 4th guided image highlighted in the red box. The middle case edits the final view with ``a single tree", and the right case edits the final view with ``lush trees".
This demonstrates the model is capable of generating panoramic images with both semantic and style controls. Please note, in order to generate a consistent image, the context view may be changed accordingly. 

\paragraph{Text Guidance}
As shown in Figure \ref{fig:control-text},
 the cases presents the results given the text guidance. In the Figure, the first row contains the original panorama images, the second row presents the text guidance and the third row illustrates the generated panoramic images. As shown in these cases, the highlighted red box shows the corresponding region modified. The semantic text guidance is ``lush trees", and the trees in these images are modified accordingly.

\section{Conclusion}
\label{sec:conclusion}
In this study, we present an innovative framework for crafting high-resolution panoramic visuals, skillfully integrating spherical structure and semantic control. By employing spherical modeling, we adeptly tackle spherical distortion and edge continuity challenges while facilitating generation through image and text cues. Future endeavors will focus on embedding interactive features and enhancing inference speed, ultimately positioning the model as a viable alternative to current human-built interfaces. 
\section{Acknowledgements}
This work was conducted during Kun Yan's internship at Microsoft Research Asia and supported in part by NSFC 61925203 and U22B2021.

\bibliography{aaai24}
\appendix
\onecolumn
\section{Implementation Details}

Our model consists of 24 transformer layers, each with 20 attention heads and a hidden size of 1280. The dropout rate for attention and feed-forward is 0.1. We update the model parameters using AdamW with $\beta_1$=0.9, $\beta_2$=0.96, $\epsilon$ = 1e-8, and a weight decay multiplier of 4.5e-2. To prevent the gradient from exploding, we clip the decompressed gradients by norm using a threshold of 4, but only during the warm-up phase at the start of training.
We trained the model using 64 NVIDIA V100 GPUs, each with 32GB memory, and a total batch size of 256, for a total of 20 epochs, corresponding to approximately 4,000 steps. To initialize the learning process, we use a linear schedule to ramp up the learning rate to 1e-4 over 5000 updates, followed by a cosine schedule that reduces the learning rate to 0.
\section{Dataset Diversity}
We incorporating the Matterport3D\cite{Matterport3D} indoor scene dataset to thoroughly assess our method's performance. In Table 1, by benchmarking our method against prominent approaches, we have showcased its generalizability and resilience in diverse data distributions. 

\begin{table}[h]
\centering
\begin{tabular}{l|c|c|c}
\hline
Method                          & FID $\downarrow$ & S-FID $\downarrow$ & Continuity \\
                                &                 & mean                      & LRCS $\downarrow$ \\
\hline
COCO-GAN                  & 87.92          & 96.36                     & 0.0867 \\
InfinityGAN                           & 81.43          & 89.17                     & 0.1029 \\
Text2light& 33.96          & 50.56                     & 0.0232 \\
\hline
ARM                             & 24.86          & 40.15                     & 0.0279 \\
LPM                             & 46.33          & 56.98                     & 0.2087 \\
SPM                    & \textbf{8.26}  & \textbf{20.63}            & \textbf{0.0021} \\
\hline
\end{tabular}
\caption{Quantitative comparison of our method, COCO-GAN, InfinityGAN, Text2light, ARM, LPM, and SPM(+SRE+SC) evaluated using the Matterport3D dataset with resolution of 2048x1024}
\end{table}

\begin{figure*}[h]
\centering
\includegraphics[clip, trim=0cm 3cm 0cm 8cm,width=\linewidth]{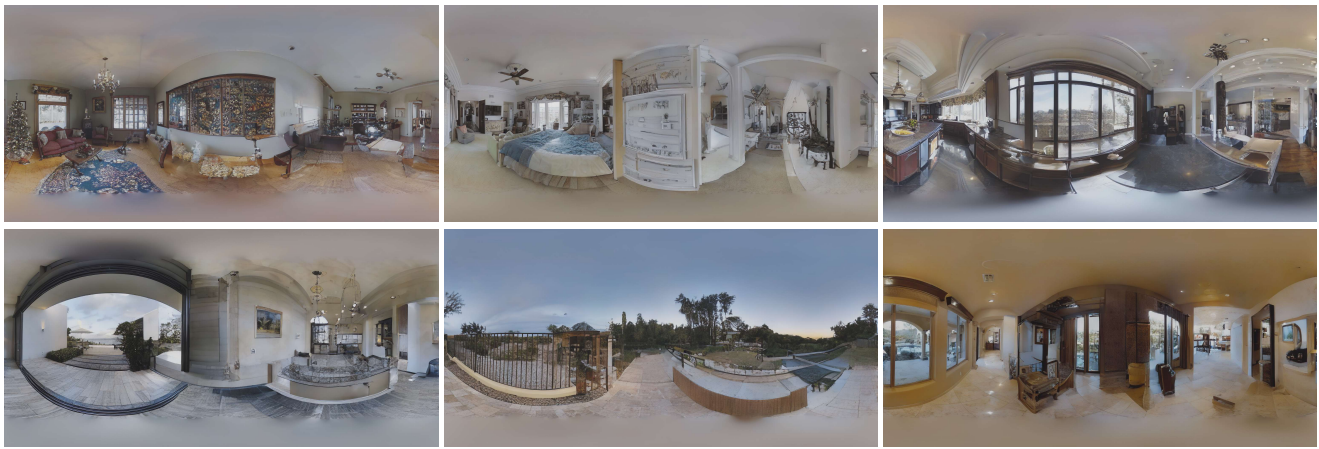} 
\caption{Samples on Matterport 3D}
\label{fig:my_label}
\end{figure*}

\section{Efficiency Ablations}
While a direct comparison with BlockadeLab's solution is not possible, we have included an estimation of their efficiency using publicly available models in Table 2 (LDM lines, represent Stable Diffusion base + 4x upscale model). The table shows our proposed method achieves better inference speed and fewer parameters compared to the LDM approach, indicating superior efficiency.
\begin{table}[h]
\centering
\begin{tabular}{l|c|c|c}
\hline
Method                 & Inference Speed           & Img/s & Parameters \\ \hline
LDM type            &$\sim$2x & 0.031 &  1228M+503M\\
ARM              & 1x            &   0.014             &   281M       \\ 
LPM              & $\sim$28x            &   0.396              &  281M             \\ 
SPM               & $\sim$14x            & 0.195                 &  297M           \\ 
\hline
\end{tabular}
\caption{Compute efficiency comparison of different methods, evaluate 1000 samples on a single A100 40GB PCIE GPU with a resolution of 2048x1024}
\end{table}

\section{Semantic Control Samples}
\begin{figure}[h!]
    \centering
    
    \subfloat[]{
    \centering
        \includegraphics[clip, trim=5.5cm 4cm 5cm 4cm, width=0.48\textwidth]{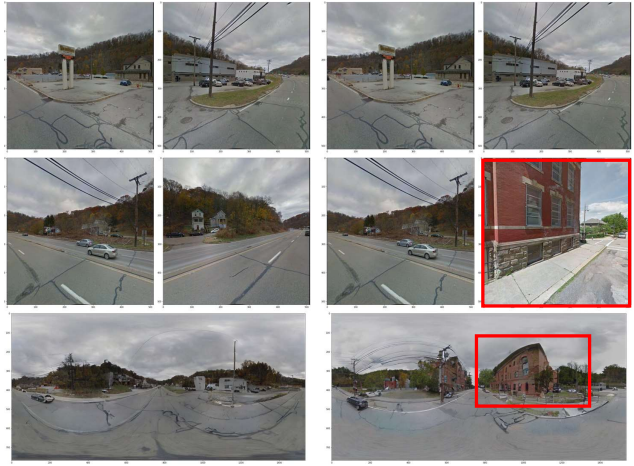}
        }
        
    \subfloat[]{
    \centering
        \includegraphics[clip, trim=5.5cm 4cm 5.5cm 4cm, width=0.48\textwidth]{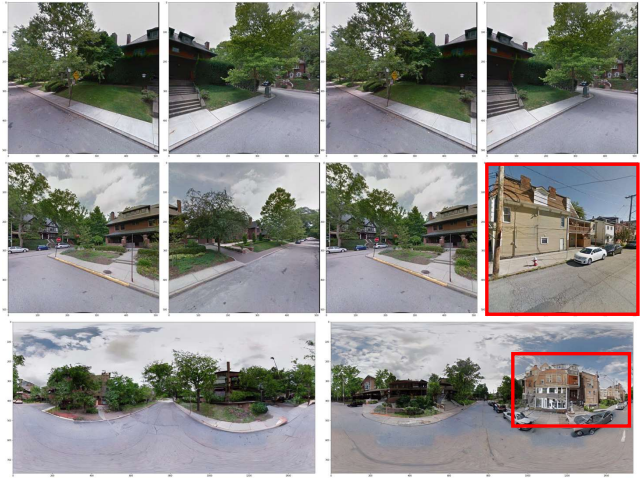}
    }
    \caption{ Visual Guided generation. When we replace the guidance with different visual semantics as shown in the middle and right columns, we can manipulate the generated panoramas as we need.}
    \label{fig:control-app}
\end{figure}

By replacing the visual conditions, our model can be used to control the semantics of the generated panoramas. As illustrated in Figure \ref{fig:control-app}, each sub-figure shows four images captured from a real view, which are then encoded as semantic conditions using CLIP and fed into our model. When we replace part of the conditions, we observe that the model adjusts the output to fit the new semantic needs (as highlighted in red boxes). This allows us to manipulate and control the generated panoramas to suit our specific needs.

\section{Comparision with similar works}
Previous studies on panorama synthesis have limitations in handling high-resolution images, particularly those with dimensions of 768x1,536. Among the relevant works that can be used for comparison are \cite{chen2022text2light,Omnidreamer}. It is important to note, however, that the datasets used in these papers are not publicly accessible. To address this issue, we compare our condition generation results with the recent state-of-the-art text2light model \cite{chen2022text2light} on the publicly available StreetLearn dataset. In addition, we evaluate the extrapolation task results of OminiDreamer \cite{Omnidreamer}.

\subsection{Comparasion with Text2light}

\paragraph{Difference on Semantic Guidance }
Our proposed method has several advantages over the text2light approach. Firstly, our method can generate robust outputs without the need for pre-computed vectors during inference, making it more flexible and adaptable to various text prompts. In contrast, the text2light method requires retrieval of embeddings from pre-computed vectors to generate plausible outputs.

Secondly, the text2light method is limited in its ability to guide the global content of the generated image, as it only provides relatively simple descriptions. On the other hand, our method can guide specific views of the panoramic image using comprehensive descriptions encoded as semantic conditions, resulting in a generation that closely aligns with the input.

In conclusion, our proposed method represents a significant improvement over the text2light approach, providing a more flexible, versatile, and comprehensive approach to panoramic image generation guided by text descriptions.

\paragraph{Comparison Protocol}
Since the HDR360-UHD used in text2light is not publicly available, we retrain the text2light model on the StreetLearn dataset, strictly adhering to the procedure and hyperparameters outlined in the original publication. The comparison is based on an embedding-conditioned setting. In our experiment, text2light utilized the global image clip embedding as the conditioning signal, while our model employed the embeddings of four different views. All results were evaluated from the test set consisting of 5.8k data points. 
 This setup was deemed reasonable as the text condition embedding used in text2light was derived from image embedding, and the global image embedding contains more information compared to just four patch views, thus providing a stronger basis for comparison.

\paragraph{Resolution Comparison}
Although the evaluation of our work is currently limited to the StreetLearn dataset, our method has already surpassed the first stage of the text2light approach in terms of resolution. This is achieved through our advanced conditioning design, which is capable of higher resolution without relying on superresolution techniques. In contrast, text2light's autoregressive design has inherent limitations in resolution, and superresolution techniques cannot add semantic detail to the generated images, as our method can.
Furthermore, we demonstrate that our method can directly apply to an even higher resolution of 1536x3072, as shown in Figure 5-14. We believe that our approach has great potential for applications in higher resolution panoramic image generation.

\subsection{Comparasion with OminiDreamer}
\begin{table}[h]
\centering
\label{tab:comparison}
\vspace{-12pt}
\begin{tabular}{ccc}
\toprule
Method & Ours &  \cite{Omnidreamer}\\
\midrule
FID & 6.93 & 10.06 \\
\bottomrule
\end{tabular}
\vspace{-5pt}
\caption{Comparison between our method and \cite{Omnidreamer} on streetlearn.}
\label{tab}
\vspace{-12pt}
\end{table}

\paragraph{Experiment Setting}
We compared our results with OmniDreamer \cite{Omnidreamer}, which mainly focuses on the extrapolation task. However, as the SUN360 dataset used in their paper is no longer available, we trained the OmniDreamer model on the StreetLearn dataset. We reported the results using the FID score, with a resolution of 1024x512 and a $180^{\circ}\times90^{\circ}$ input, following the process described in \cite{Omnidreamer}.
The comparison results, presented in Table \ref{tab}, demonstrate that our model outperforms the OmniDreamer approach.

\paragraph{Method Comparison}\cite{Omnidreamer} employs circular inference to address the issue of in-continuity. Although it can deal with coarse continuity, some minor semantic inaccuracies still persist due to the direct replacement of the latent code. We instead address this problem by gradually refining the content during the training phase, avoiding this problem at an early stage. In the revised version of the manuscript, we will provide more qualitative examples to further illustrate these differences. Besides this advantage,
We also want to emphasize that our method offers several advantages over \cite{Omnidreamer}. Our approach supports higher resolution extrapolation and has additional controllability on the completion region, which eliminates the limitations discussed in \cite{Omnidreamer}.

\section{Assumptions and Approach}

\paragraph{Horizontal Assumption}
The horizontal assumption is prevalent in panoramic image capture, as most panoramas are captured horizontally and from outdoor environments. Our method is naturally well-suited for this type of image capture and can handle slight misalignments due to non-trimmed top and bottom latent code quantity. This results in robust performance and ensures that our method can generate high-quality panoramic images even with imperfect input data.

\paragraph{Why SPM after LPM} We find directly applying SC on LPM can not deliver strong results. We interpret it as the LPM and SPM steps actually performing coarse generation and refinement sequentially, and lacking of each will cause a performance drop. Besides, spherical refinement based on early noise may cause the training unstable. 
\section{Limitations of HORIZON} We observe some artifacts may result from the limited capacity of VQGAN. We will include a discussion on this aspect in the revised paper. In our future work, we aim to improve the encoder design by combining global positional encoding into VQGAN. This enhancement could possibly mitigate these artifacts and improve the overall quality of the panorama generation. 

\begin{figure}[h!]
    \centering
    \subfloat[]{
    \centering
    \includegraphics[clip, trim=0.5cm 4.5cm 0.5cm 4.3cm,width=0.5\textwidth]{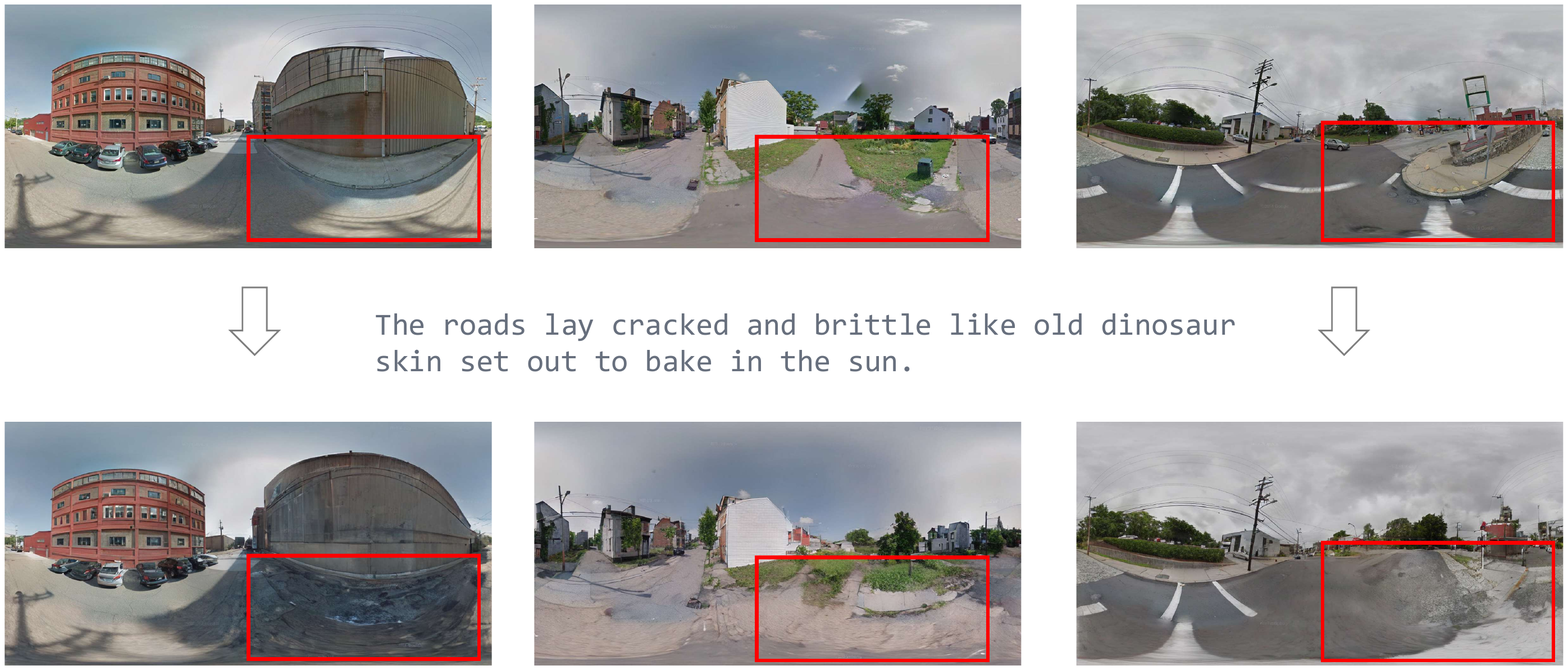}
    }
     \\
    \subfloat[]{
     \centering
     \includegraphics[clip, trim=1.8cm 4.5cm 0.5cm 4.3cm,width=0.5\textwidth]{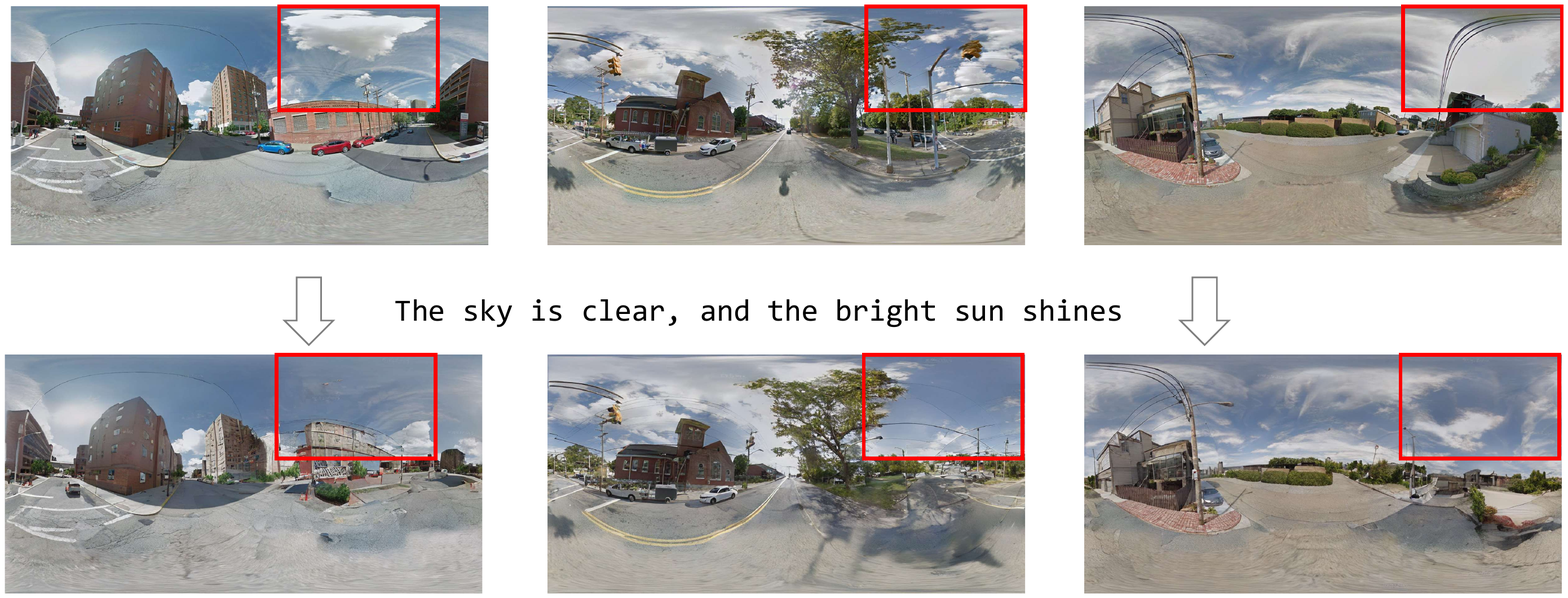}
              }
     \\
     \subfloat[]{
     \centering
    \includegraphics[clip, trim=4.2cm 4.5cm 3.5cm 4.3cm,width=0.35\textwidth]{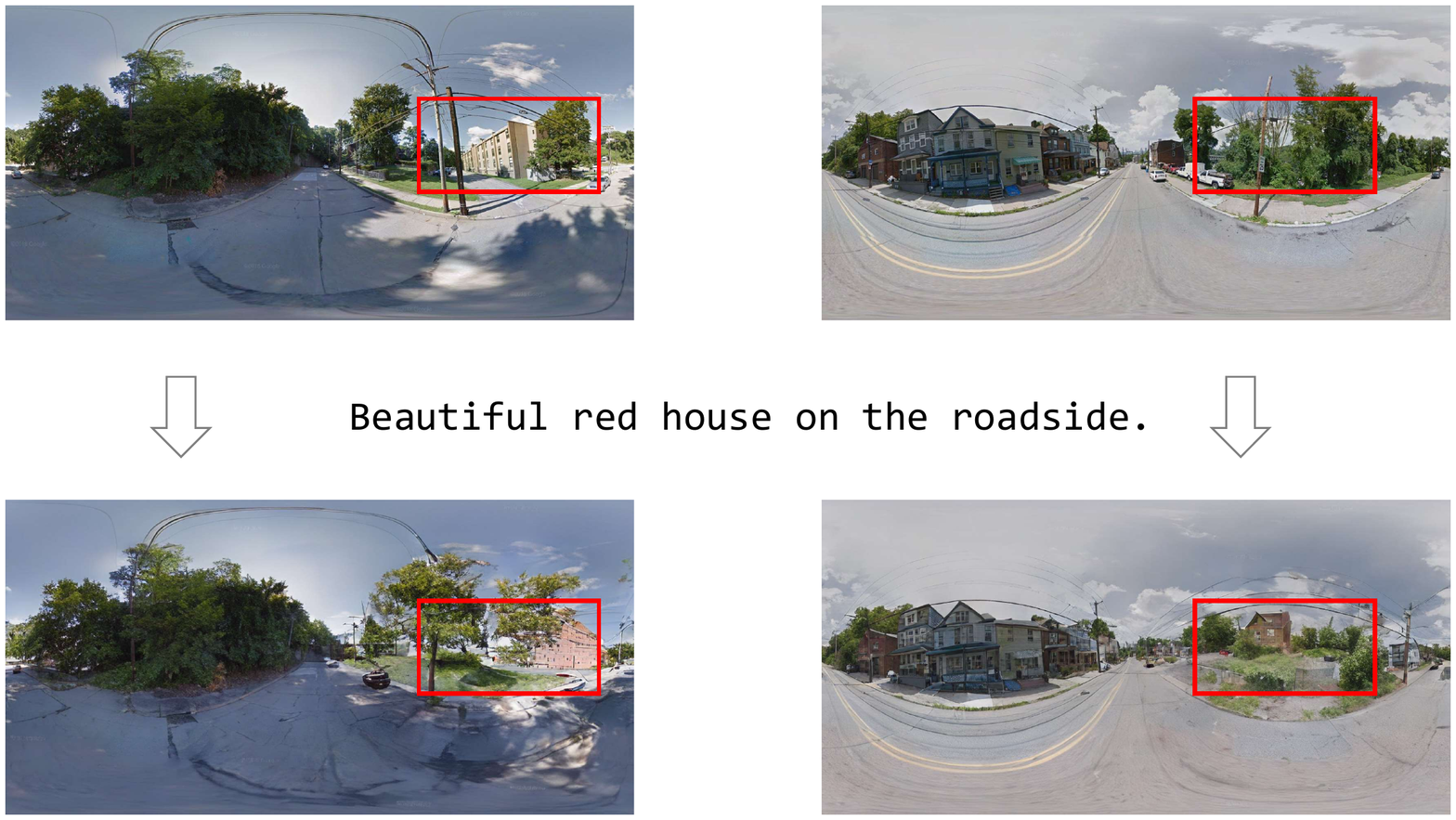}
    }
    \caption{ Textual Guided generation. We can also use natural language to edit or embellish target panoramas. In each case, the top role are the original panoramas, and the bottom role are the embellished panoramas according to the text hint shows in the middle.}
    \label{fig:control-text-app}
    \vspace{-15pt}
\end{figure}

\section{More Samples}
\begin{figure}
\centering
  \subfloat[]{
    \includegraphics[width=0.9\linewidth]{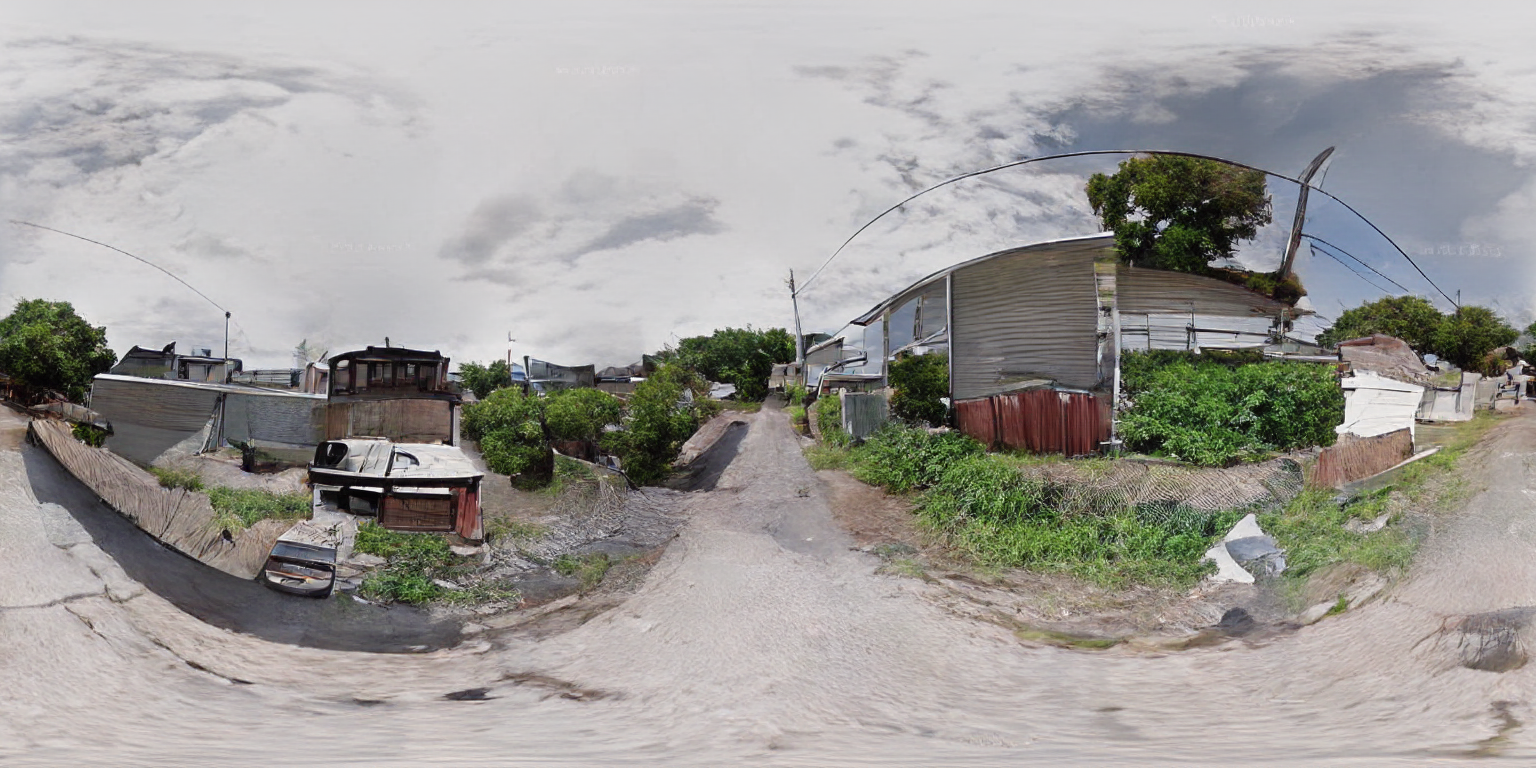}
  }\\
    \subfloat[]{
    \includegraphics[width=0.9\linewidth]{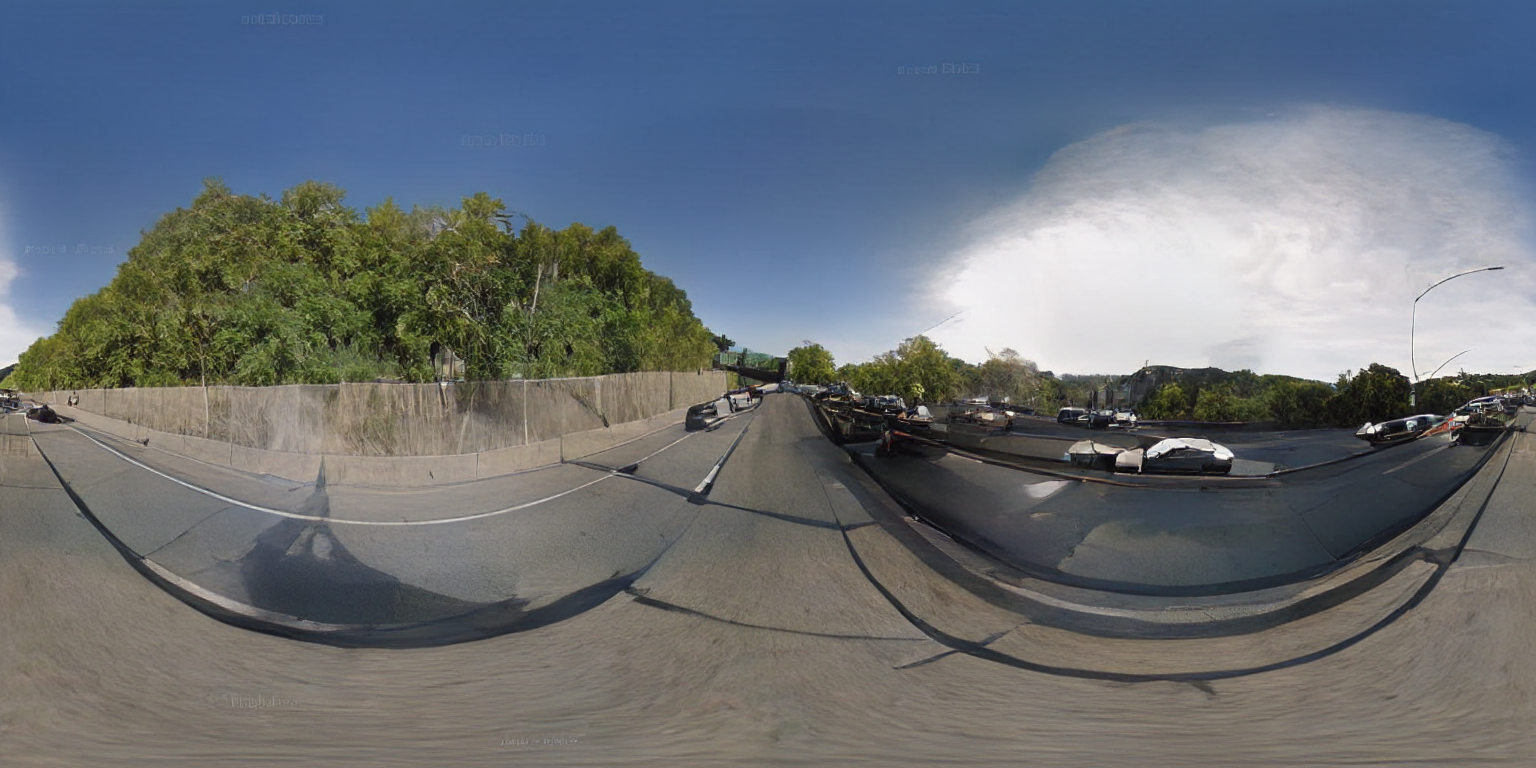}
  }\\
  \subfloat[]{
    \includegraphics[width=0.9\linewidth]{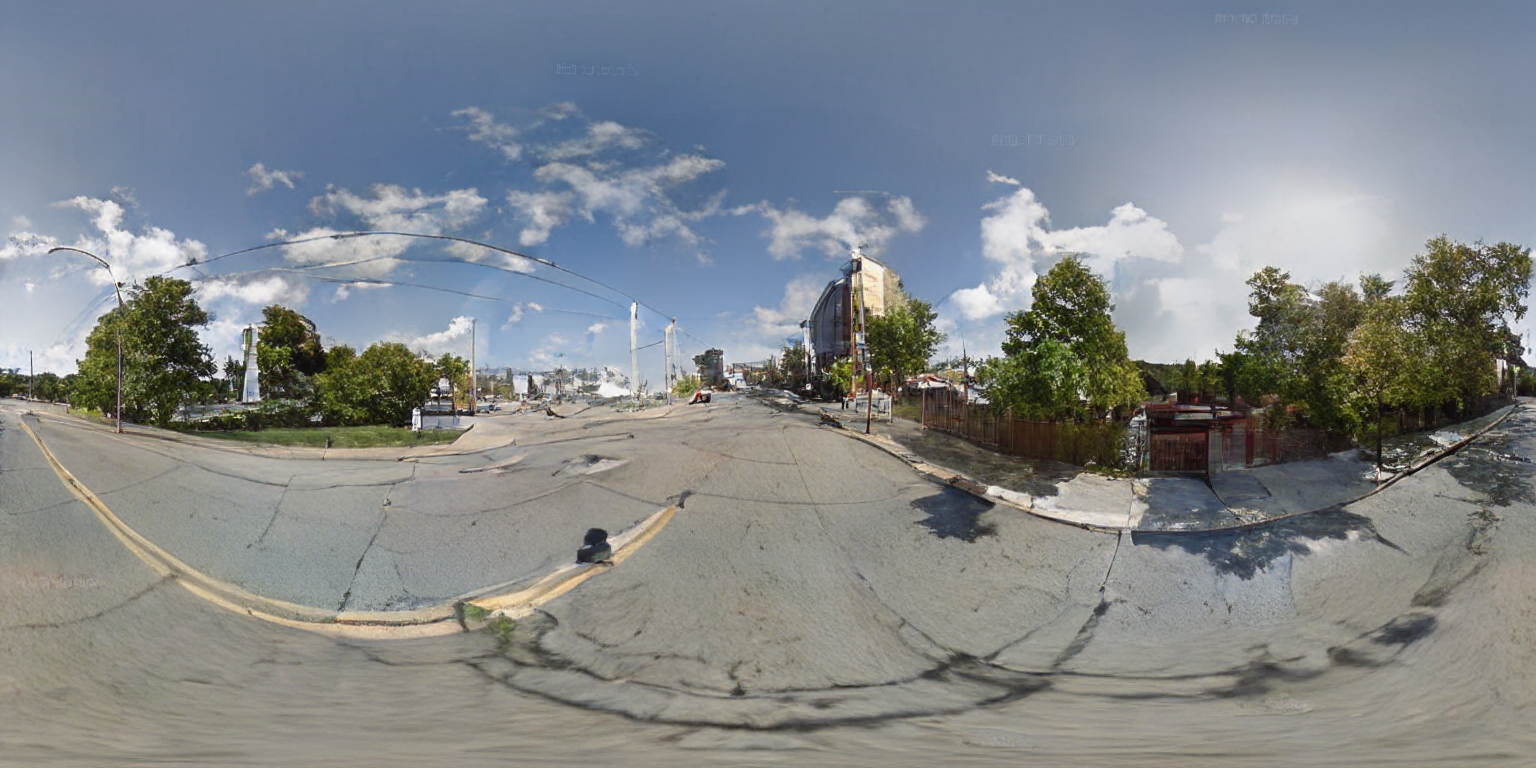}
  }\\
\end{figure}
\begin{figure*}
\centering
  \subfloat[]{
    \includegraphics[width=\linewidth]{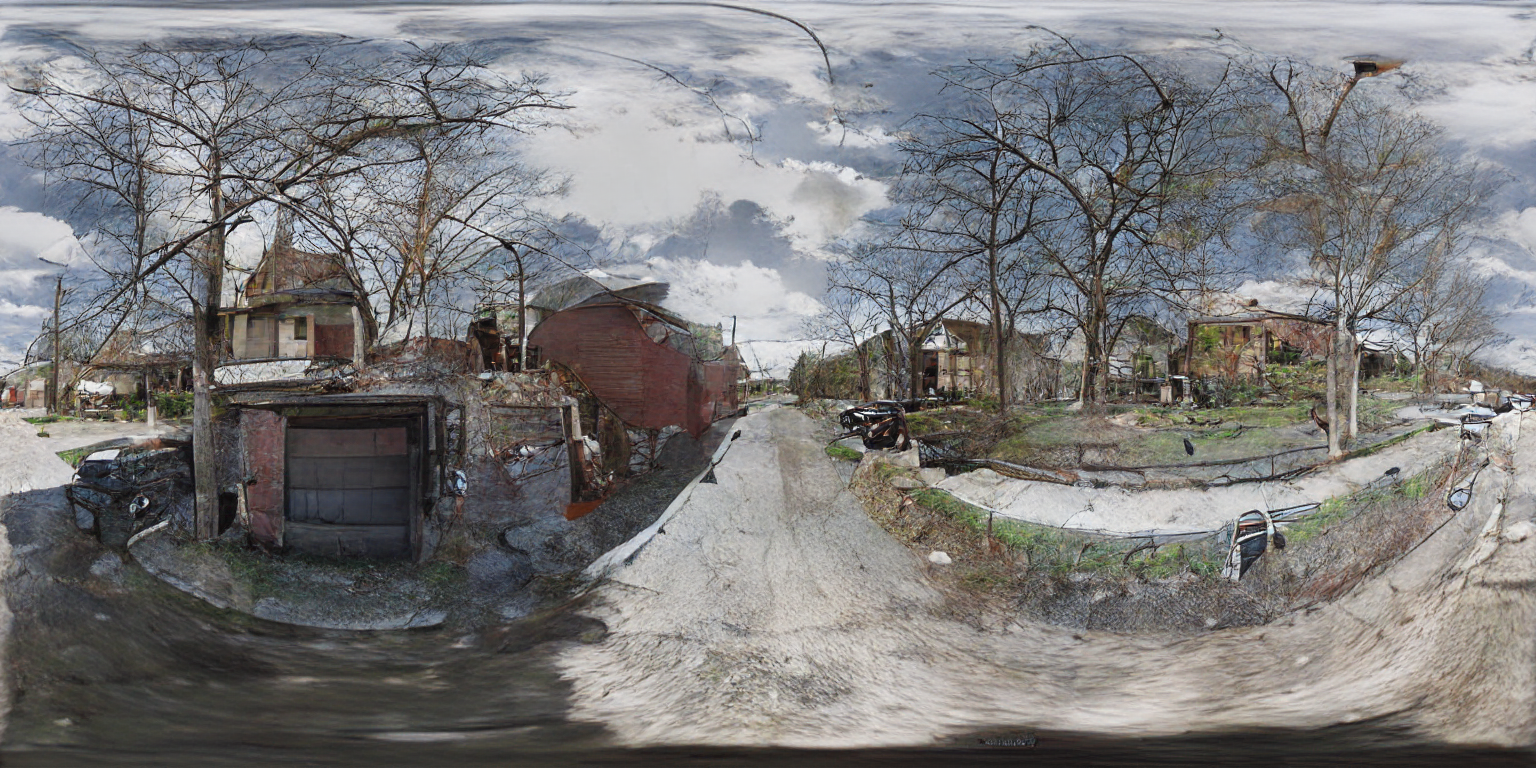}
  }\\
  \subfloat[]{
    \includegraphics[width=\linewidth]{fig/example/case15.png}
  }\\
\end{figure*}
\begin{figure*}
\ContinuedFloat
\centering
  \subfloat[]{
    \includegraphics[width=\linewidth]{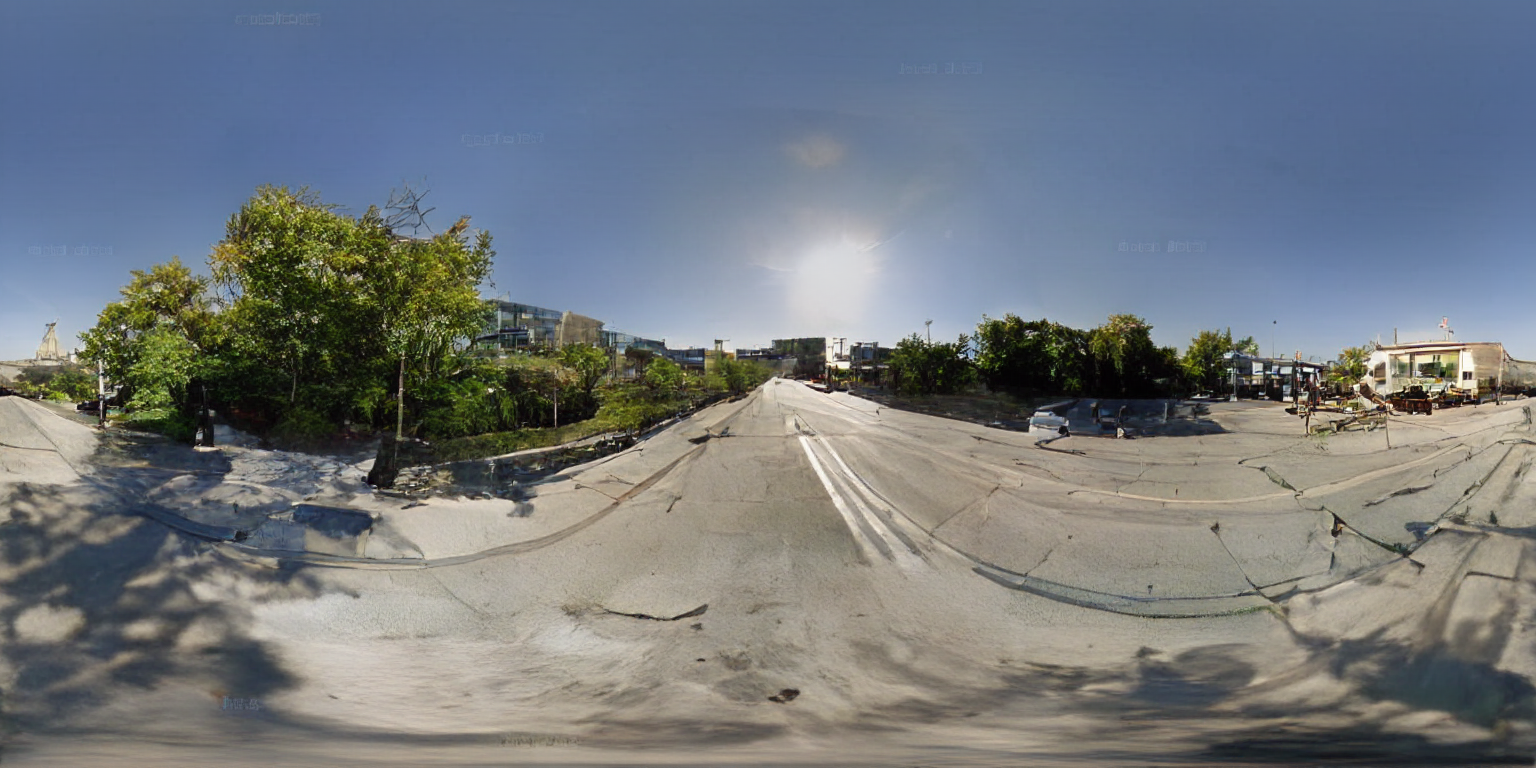}
  }\\
  \subfloat[]{
    \includegraphics[width=\linewidth]{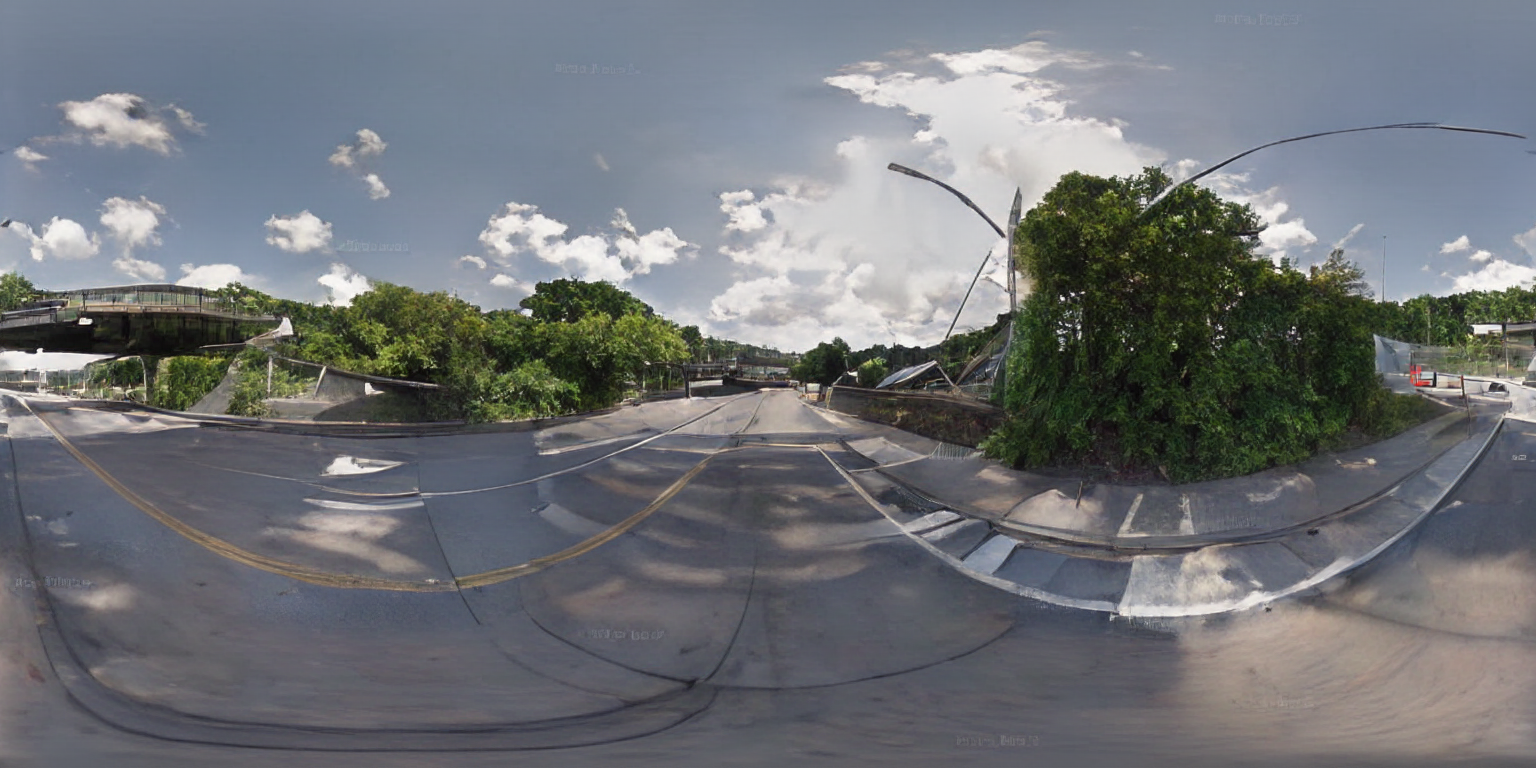}
  }\\
\end{figure*}

\begin{figure*}
\ContinuedFloat
\centering
  \subfloat[]{
    \includegraphics[width=\linewidth]{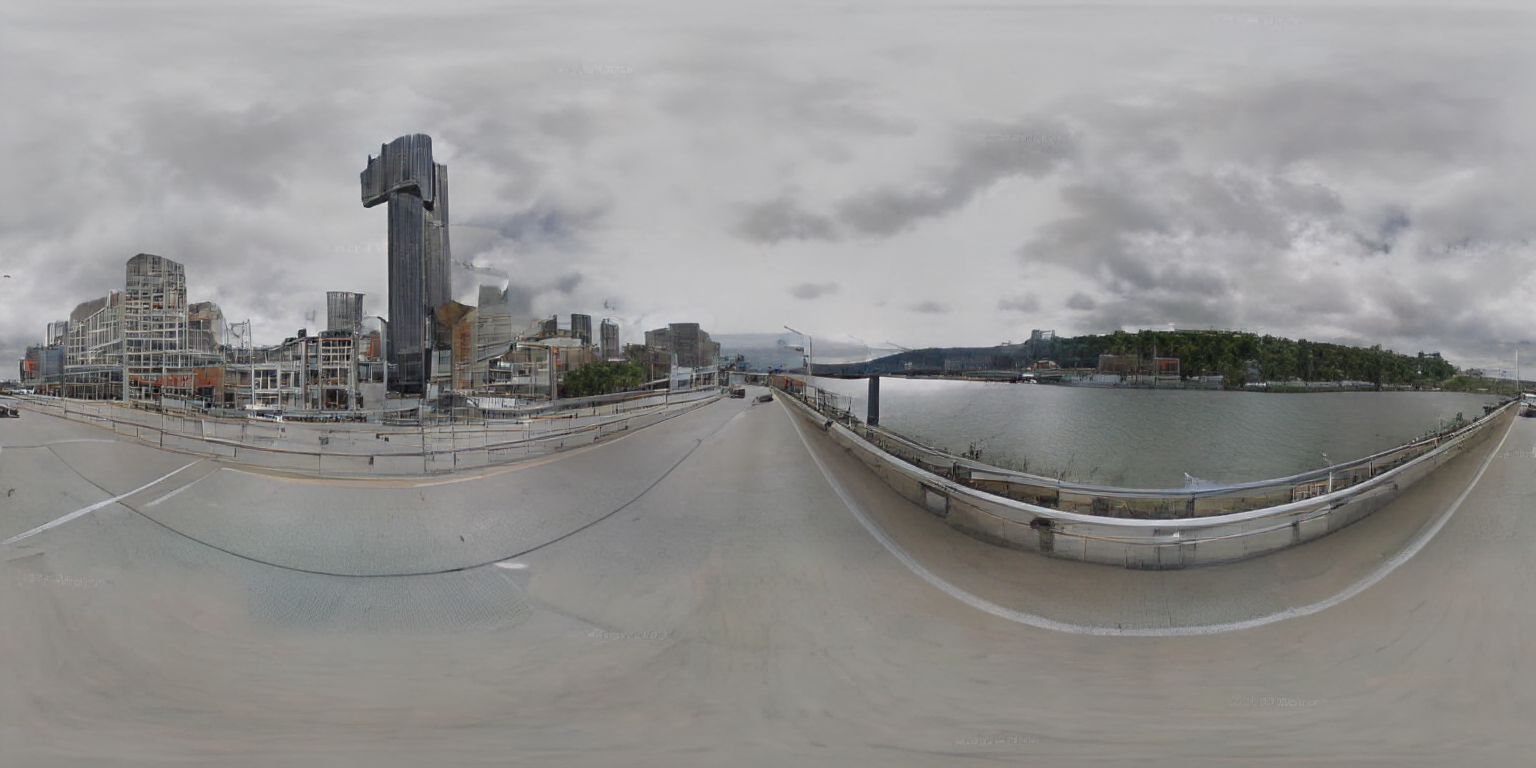}
  }\\
  \subfloat[]{
    \includegraphics[width=\linewidth]{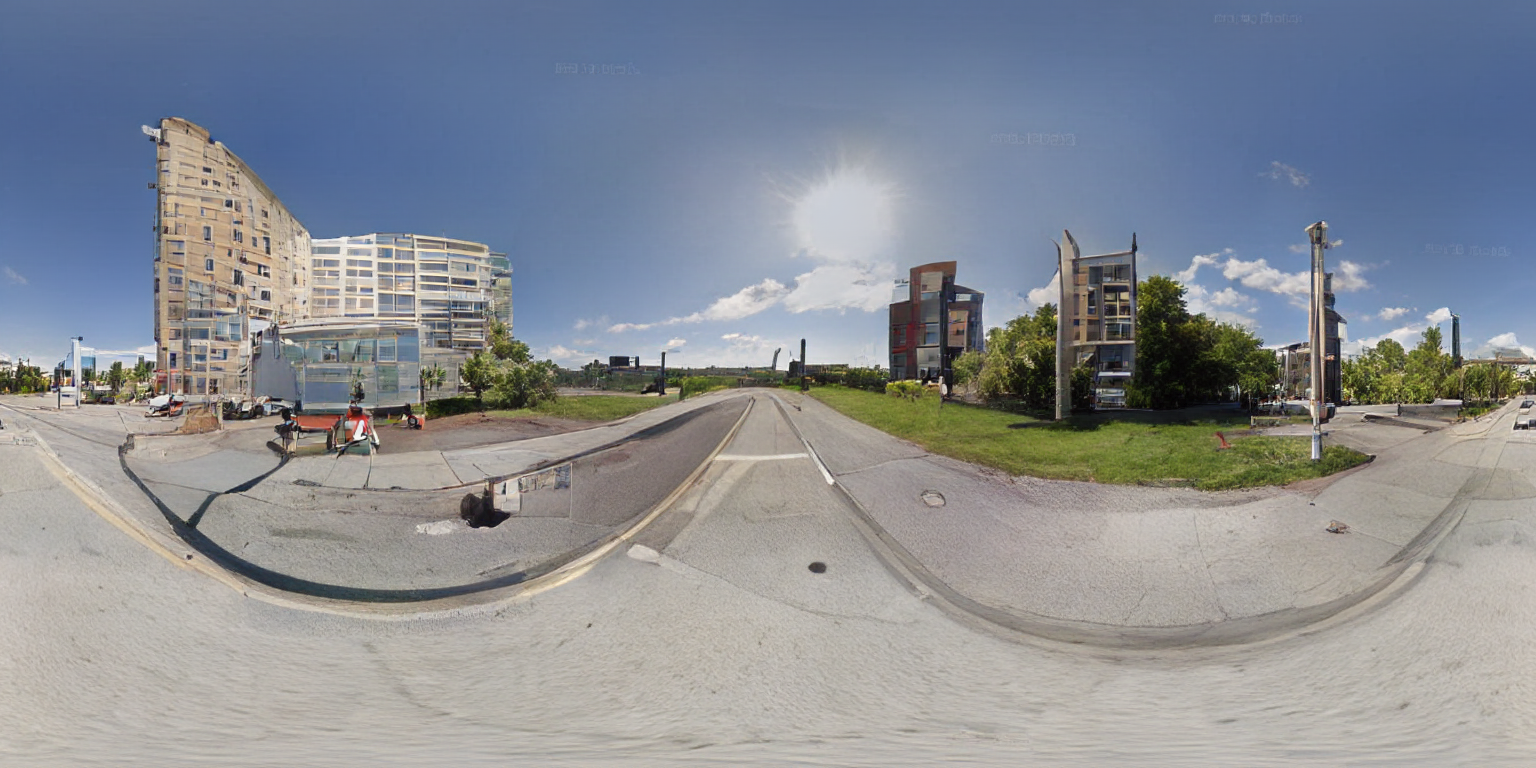}
  }\\

\end{figure*}

\begin{figure*}
\ContinuedFloat
\centering
  \subfloat[]{
    \includegraphics[width=\linewidth]{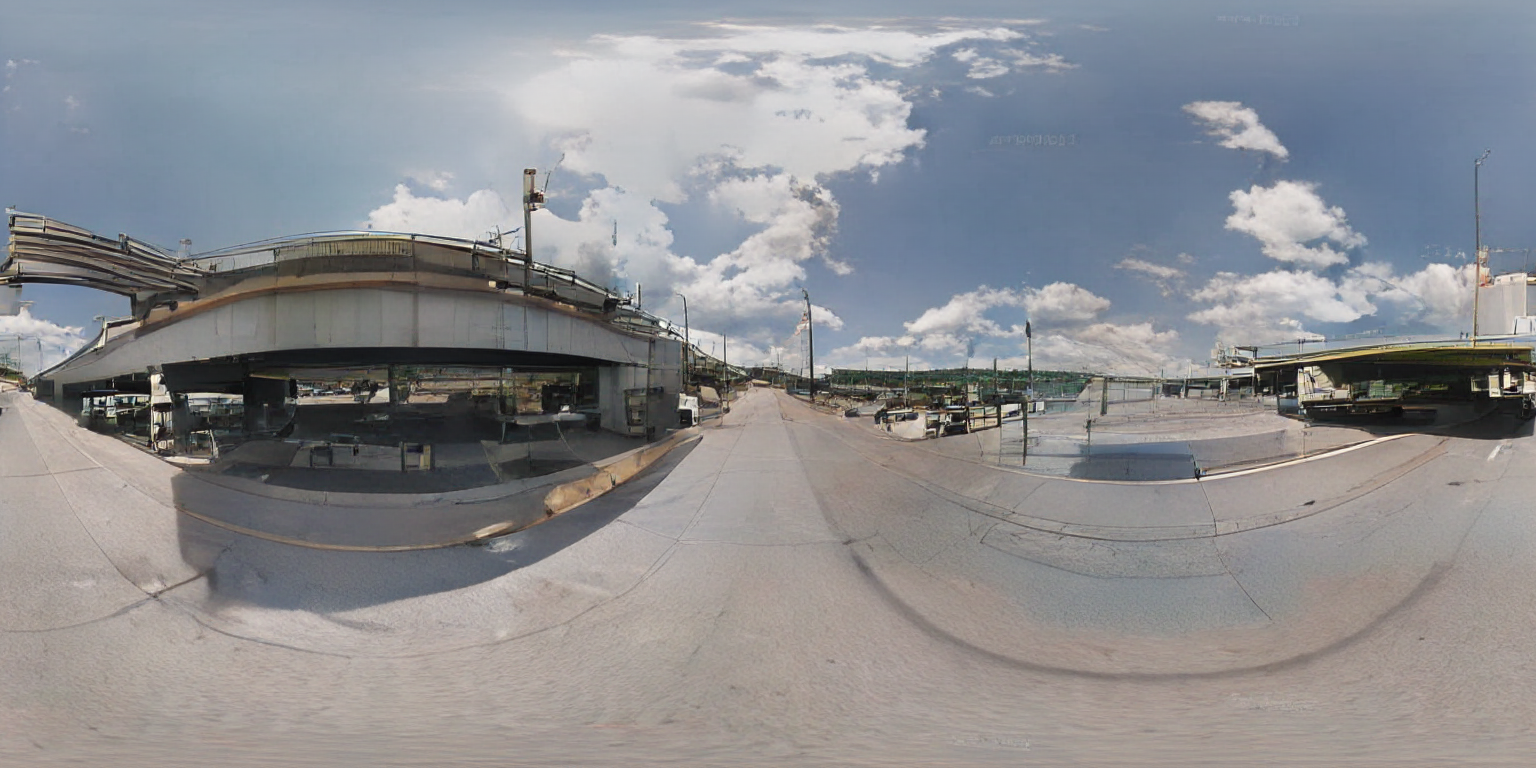}
  }\\
  \subfloat[]{
    \includegraphics[width=\linewidth]{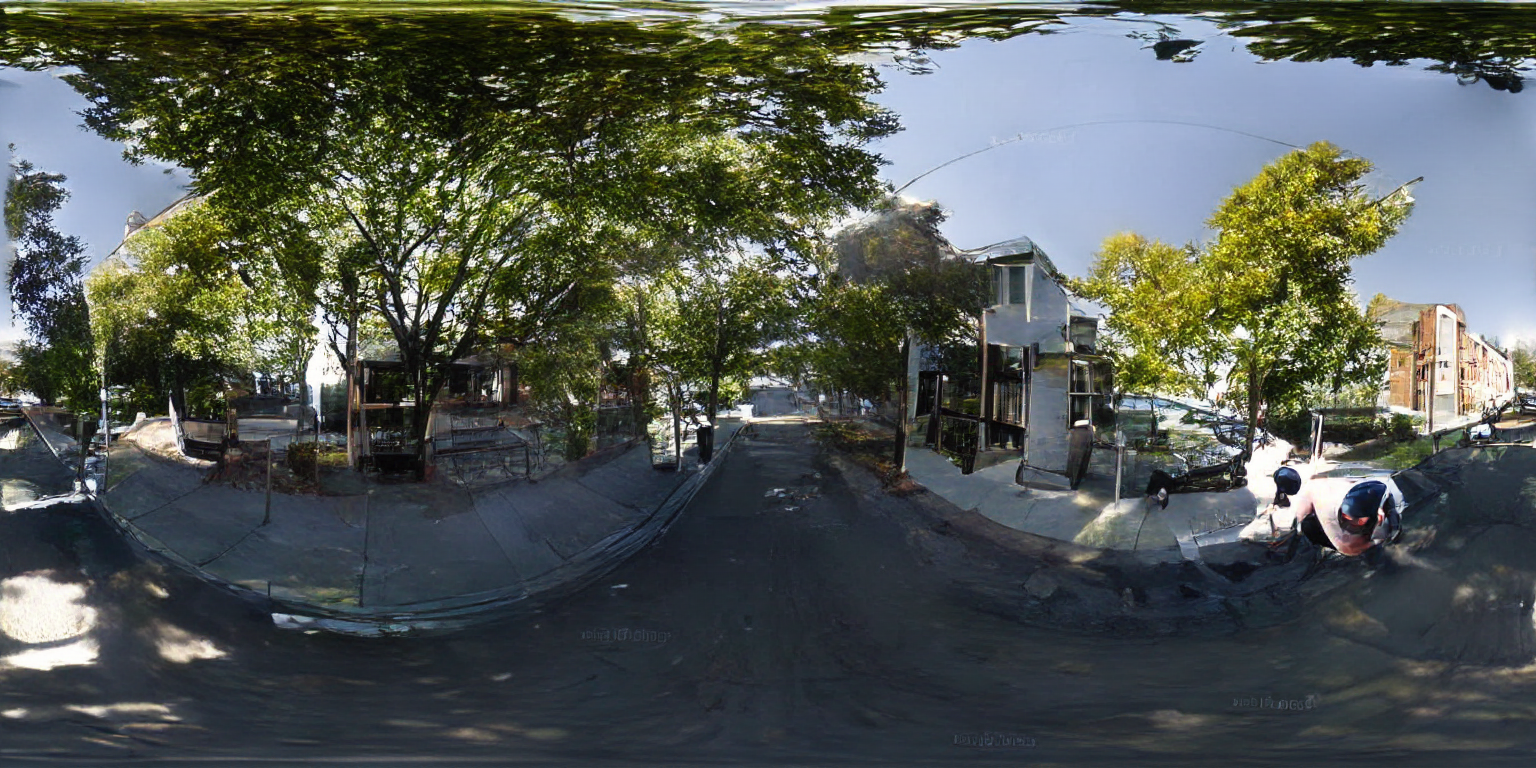}
  }\\

\end{figure*}

\begin{figure*}
\ContinuedFloat
\centering
  \subfloat[]{
    \includegraphics[width=\linewidth]{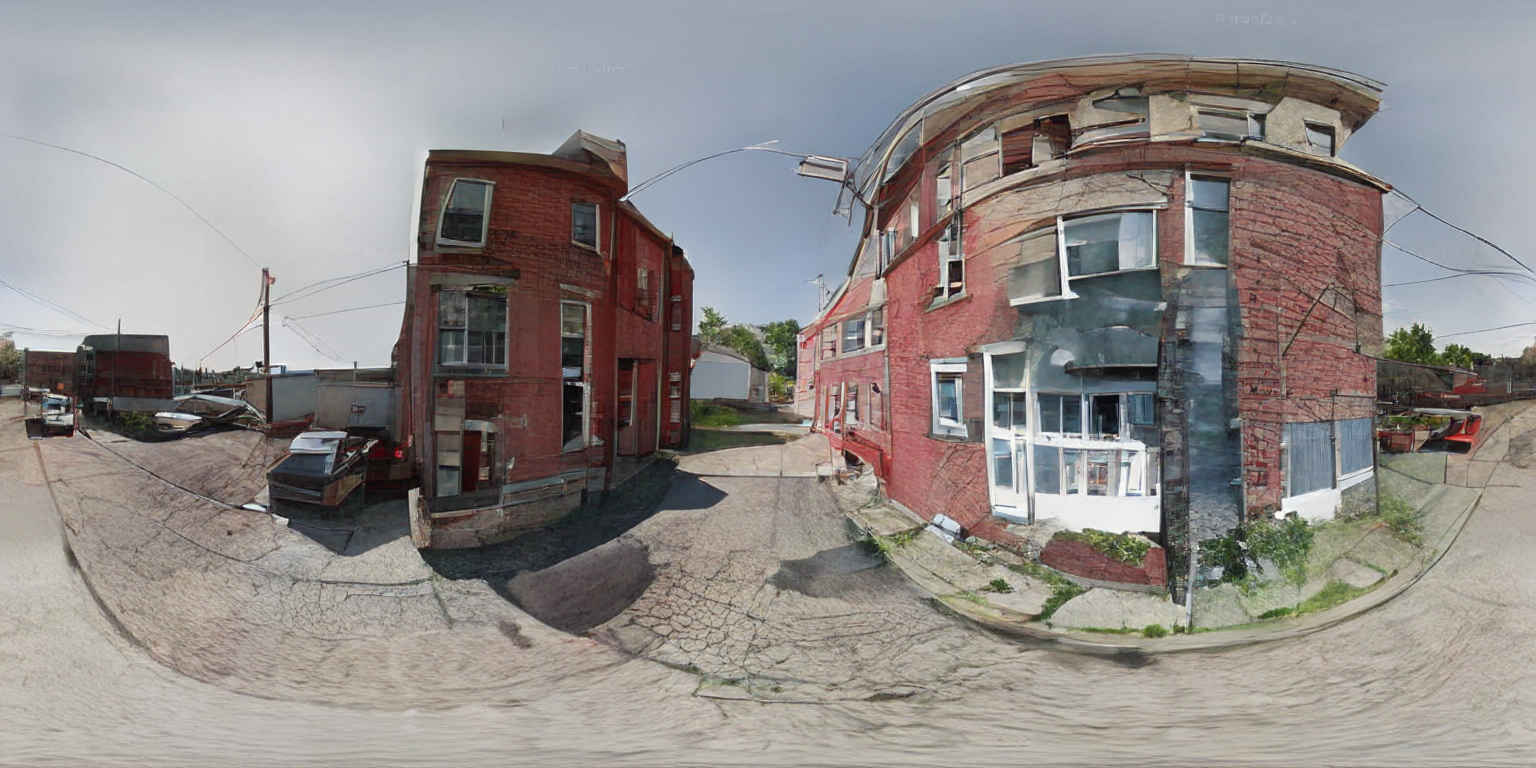}
  }\\
  \subfloat[]{
    \includegraphics[width=\linewidth]{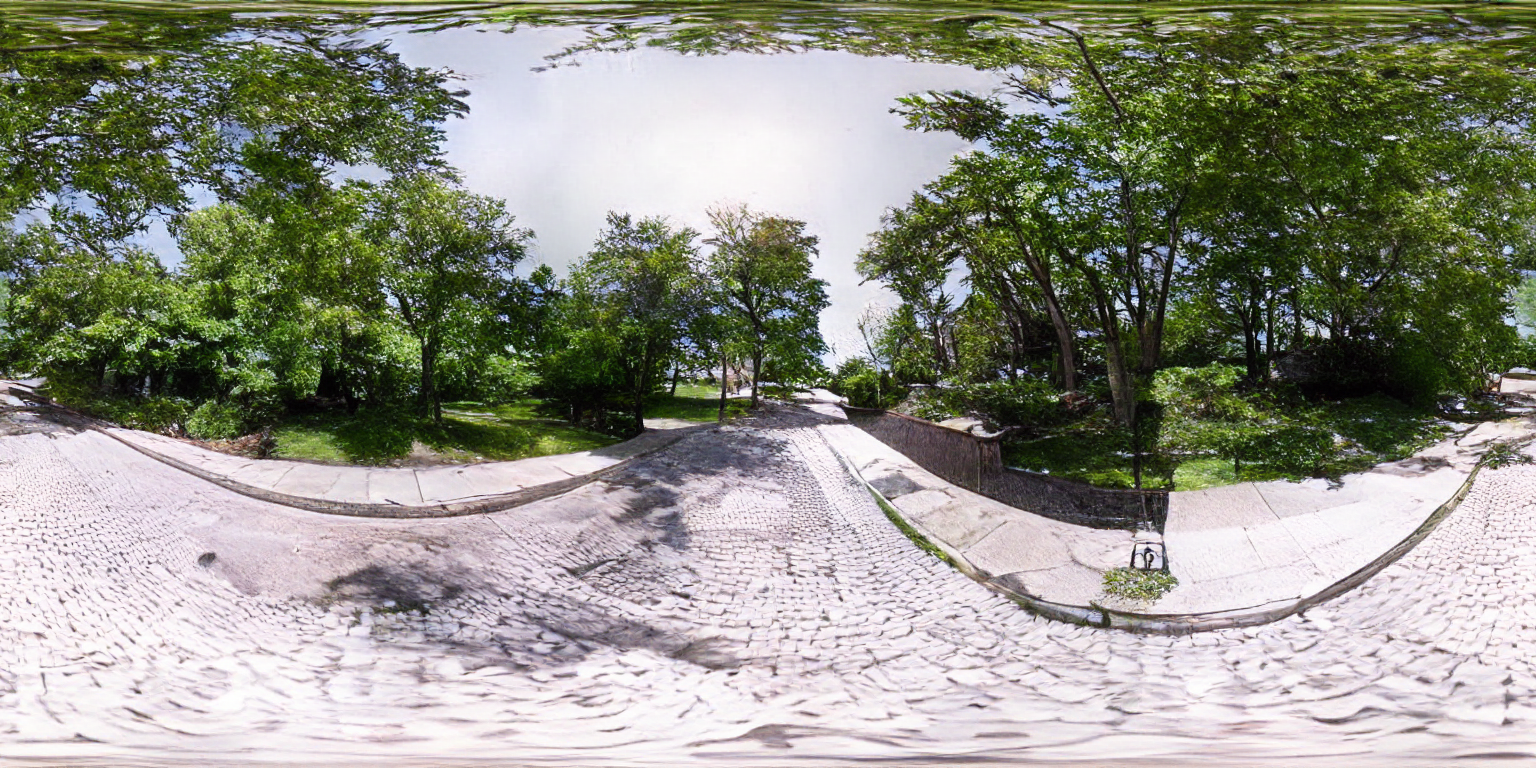}
  }\\

\end{figure*}

\begin{figure*}
\ContinuedFloat
\centering
  \subfloat[]{
    \includegraphics[width=\linewidth]{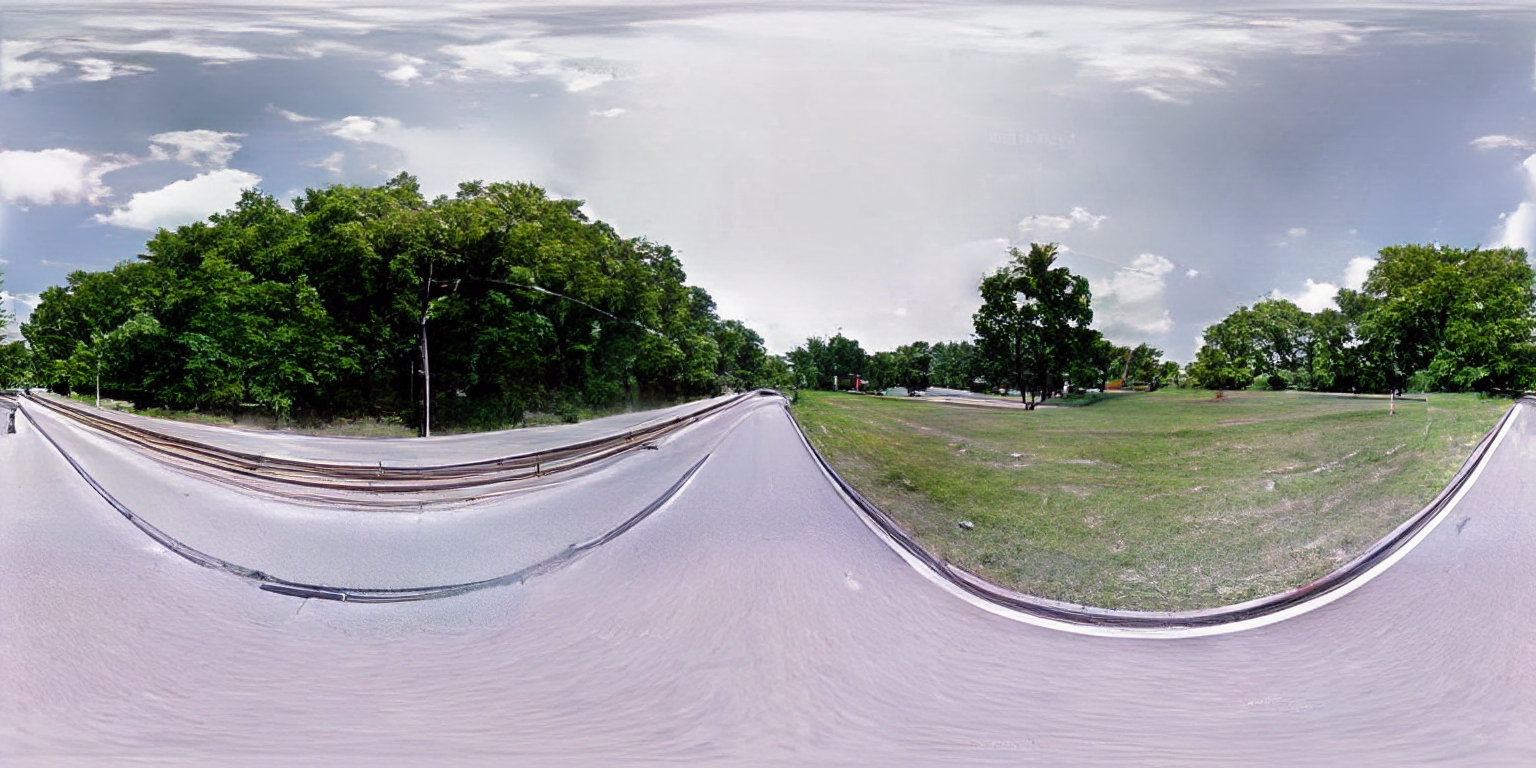}
  }\\
  \subfloat[]{
    \includegraphics[width=\linewidth]{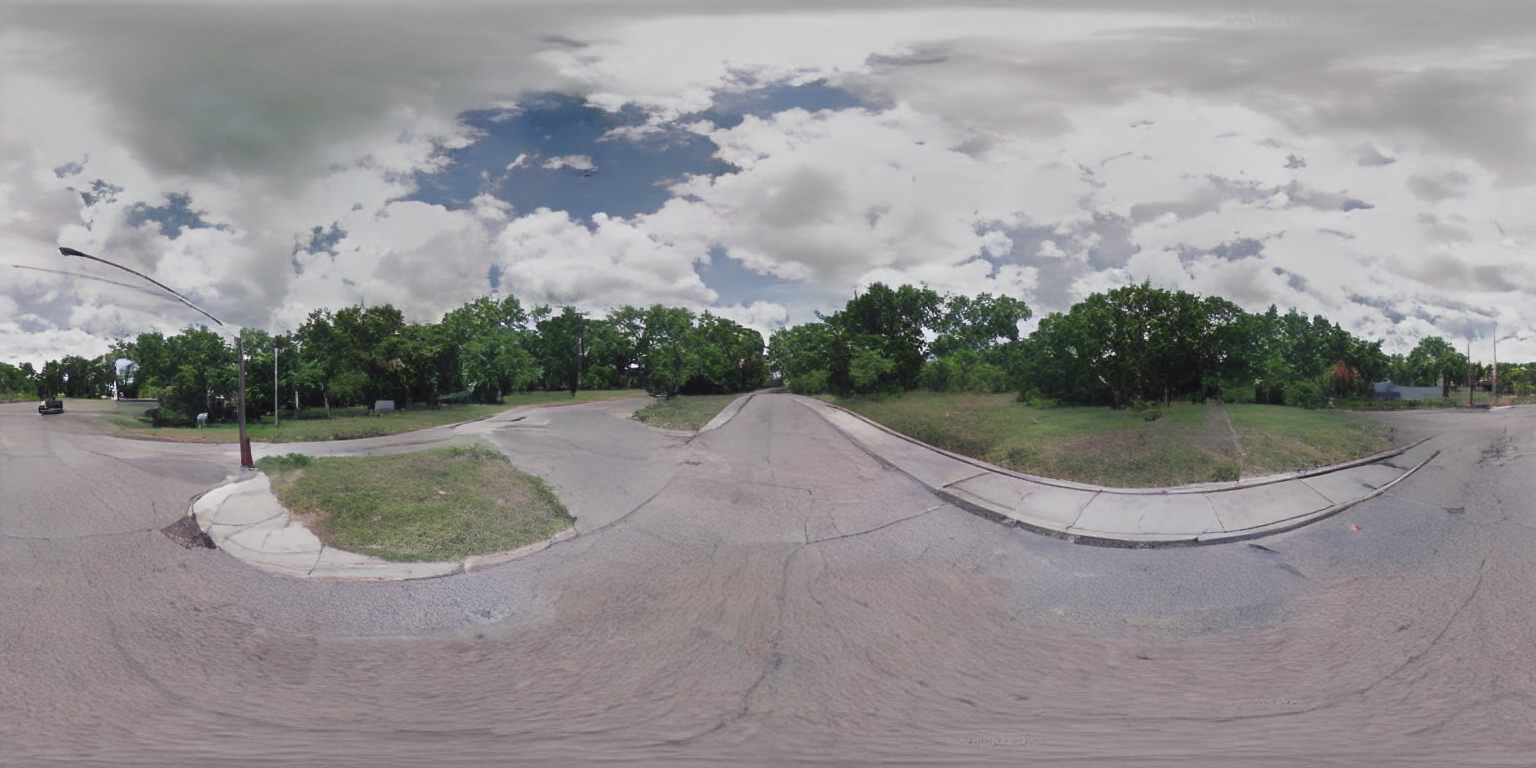}
  }\\

\end{figure*}

\begin{figure*}
\ContinuedFloat
\centering
  \subfloat[]{
    \includegraphics[width=\linewidth]{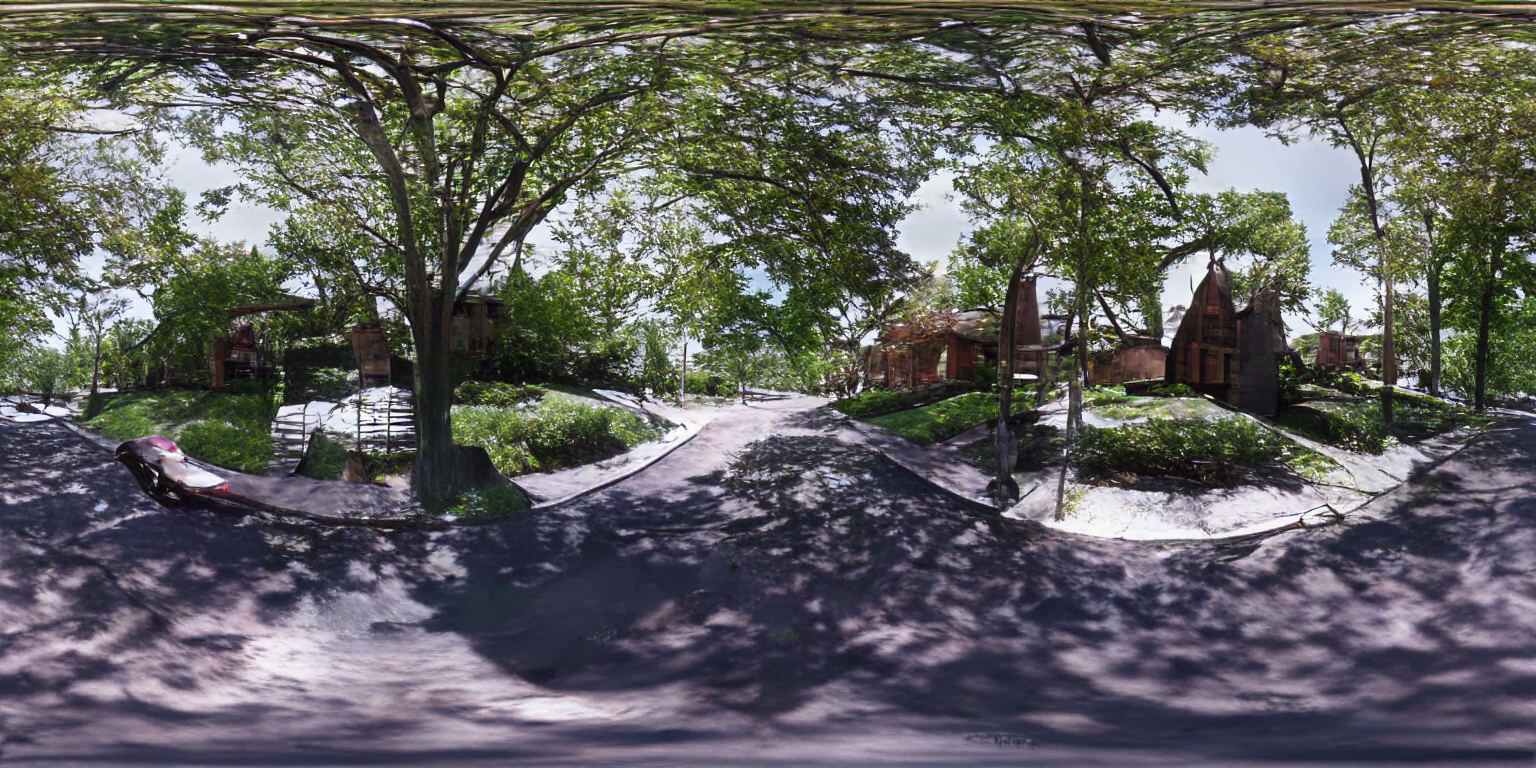}
  }\\
  \subfloat[]{
    \includegraphics[width=\linewidth]{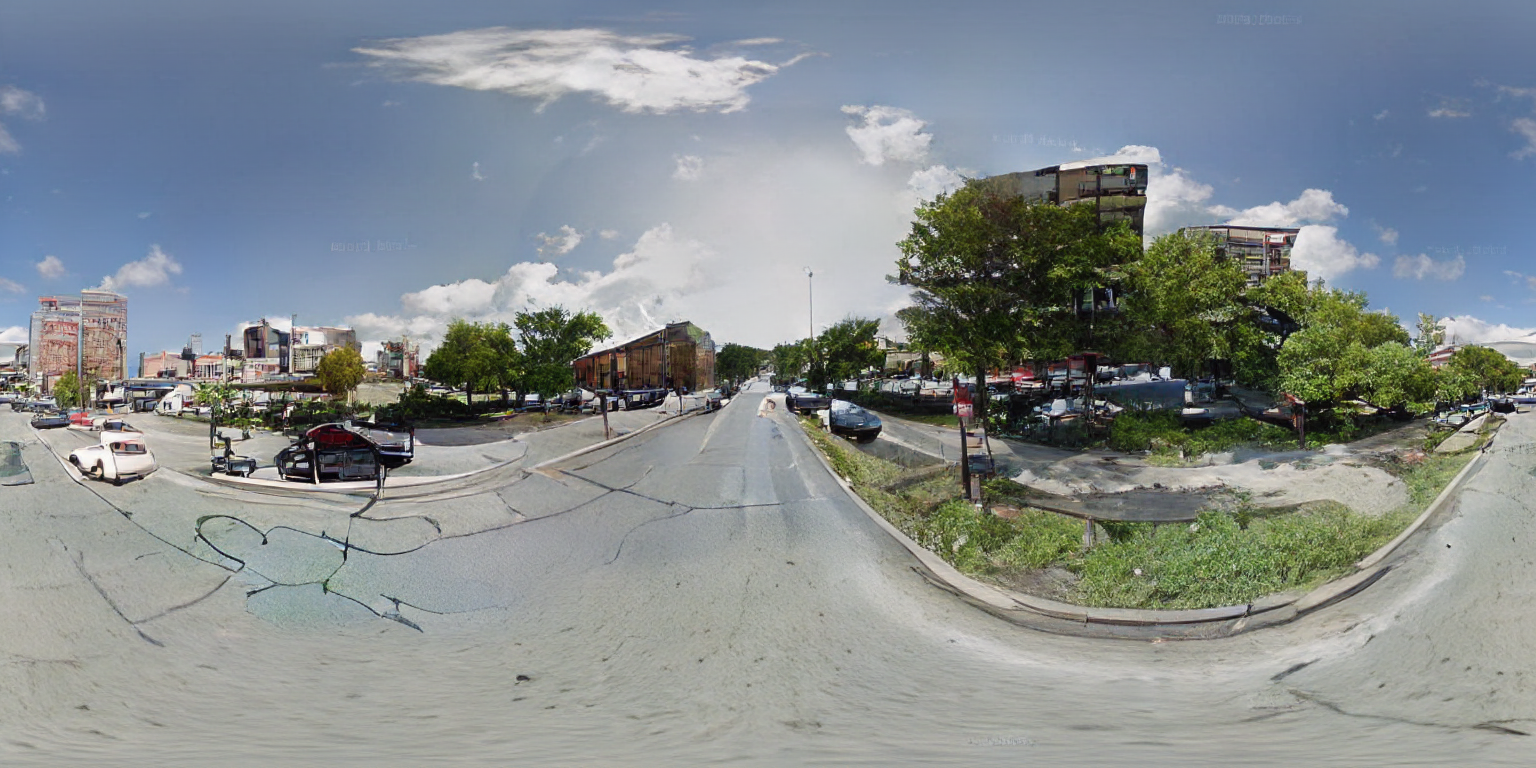}
  }\\
  \caption{More generated examples}

\end{figure*}

\end{document}